%% file: main.tex
\documentclass[10pt,twocolumn,letterpaper]{article}

\usepackage[pagenumbers]{wacv} 

\usepackage{amsmath}
\usepackage{amsfonts}
\usepackage{mathtools}
\usepackage{multirow}
\usepackage{booktabs}
\usepackage{physics}
\usepackage{todonotes}
\usepackage{makecell} 
\usepackage{enumitem}
\usepackage[table]{xcolor}
\usepackage{float}


\definecolor{wacvblue}{rgb}{0.21,0.49,0.74}
\usepackage[pagebackref,breaklinks,colorlinks,allcolors=wacvblue]{hyperref}
\usepackage[capitalize]{cleveref}

\newcommand{\cfd}[2]{#1\,{\scriptsize(#2)}}         
\newcommand{\cc}{\makebox[3.5em][c]{--}}
\newcommand{\ul}[1]{\underline{#1}}                  
\newcommand{\std}[1]{\scriptsize $\pm$#1}     
\newcommand{\stdp}[2]{\scriptsize $\pm$#1 ($\pm$#2)}     
\newcommand{\avgstd}[2]{#1\,{\scriptsize$\pm#2$}}   

\def\wacvPaperID{2536} 
\def\confName{WACV}
\def\confYear{2026}

\title{\textsc{FNOpt}: Resolution‑Agnostic, Self‑Supervised Cloth Simulation using Meta‑Optimization with Fourier Neural Operators}

\author{
Ruochen Chen\thanks{Equal contribution}
\quad
Thuy Tran\footnotemark[1]
\quad
Shaifali Parashar\vspace{.15cm}\\
\small CNRS, École Centrale de Lyon,
INSA Lyon,
Université Claude Bernard Lyon 1, 
LIRIS, UMR5205, France\\
{\tt\small \{ruochen.chen, dinh-vinh-thuy.tran, shaifali.parashar\}@liris.cnrs.fr}
}

\begin{document}
\maketitle


\begin{abstract}
We present \textsc{FNOpt}, a self-supervised cloth simulation framework that formulates time integration as an optimization problem and trains a resolution-agnostic neural optimizer parameterized by a Fourier neural operator (FNO). Prior neural simulators often rely on extensive ground truth data or sacrifice fine-scale detail, and generalize poorly across resolutions and motion patterns. In contrast, \textsc{FNOpt} learns to simulate physically plausible cloth dynamics and achieves stable and accurate rollouts across diverse mesh resolutions and motion patterns without retraining. Trained only on a coarse grid with physics-based losses, \textsc{FNOpt} generalizes to finer resolutions, capturing fine-scale wrinkles and preserving rollout stability. Extensive evaluations on a benchmark cloth simulation dataset demonstrate that \textsc{FNOpt} outperforms prior learning-based approaches in out-of-distribution settings in both accuracy and robustness. These results position FNO‑based meta‑optimization as a compelling alternative to previous neural simulators for cloth; thus reducing the need for curated data and improving cross‑resolution reliability. Our code is publicly available at \href{https://github.com/Simonhfls/FNOpt}{https://github.com/Simonhfls/FNOpt}.

\end{abstract}


\section{Introduction}\label{sec:Introduction}
Physics-based cloth simulation has been a long-standing research topic in computer graphics. Traditional approaches model the cloth dynamics by discretizing the classical time-varying partial differential equation (PDE) of motion into an ordinary differential equation (ODE), and apply various numerical integration methods for simulation. Since the cloth deformation is stiff, traditional solvers require expensive computation for either small time steps in explicit methods~\cite{breen1994predicting} or additional techniques for implicit ones~\cite{Baraff1998} in order to avoid numerical instability. This hinders real-time applications such as realistic animations in computer-aided engineering. To overcome these limitations, data-driven methods~\cite{Pfaff21, Ian2023} have emerged, where neural networks are trained on ground truth trajectories generated by traditional simulators~\cite{Narain2012}.  These neural simulators predict the next state by a single forward pass and thus significantly improve the efficiency. However, they require large amounts of curated training data, which can be costly to generate. Moreover, these models often struggle to generalize beyond the distribution of motions and mesh resolutions seen during training. 

\begin{figure}[ht!]
  \centering
\includegraphics[width=1\linewidth]{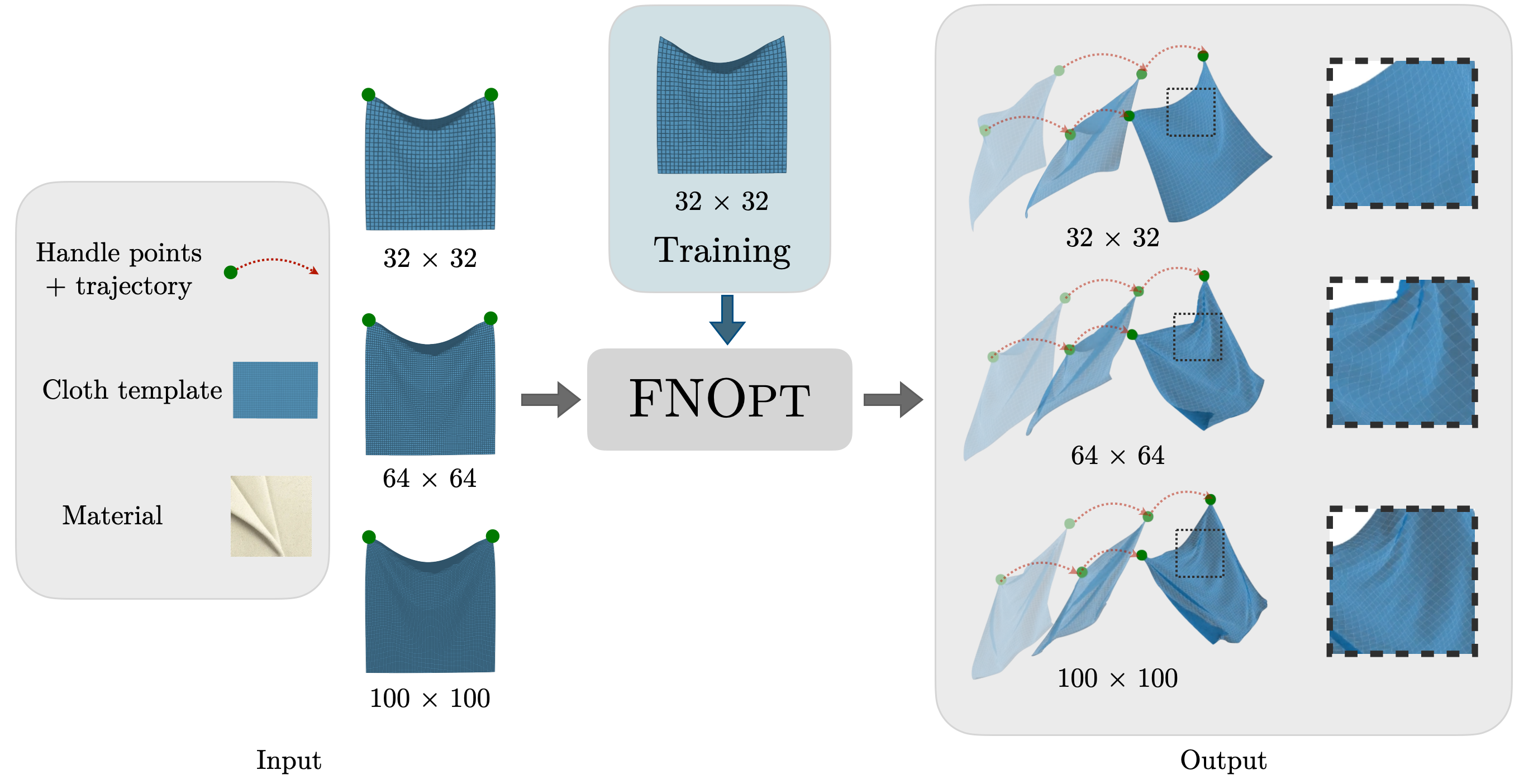}
   \caption{Being trained on $32\times32$ mesh, \textsc{FNOpt} generalizes to various mesh resolutions, cloth templates, motion speeds, handle points and trajectories without retraining.}
   \label{fig:teaaser}
\end{figure}

Modern methods~\cite{Santesteban2022, Stotko2024} recast the implicit numerical integrators for the object motion's discretized ODE as an optimization problem, which allow learning neural garment and cloth simulators by minimizing the physics-based losses. However, their rollouts often suffer from stability issues. A novel development~\cite{Nils2024} shows that rather than a neural simulator, training a neural optimizer via meta-learning technique enables adaptive iterative update that can solve various PDEs and cloth simulations at higher precision. However, while this model generalizes to various motions, it still struggles to generalize across various mesh resolutions. 
Recently, neural operators such as~\cite{Li2020, li2021} are introduced that directly learn mappings between function spaces. Unlike conventional neural networks that can only interpolate between the learned spaces, these operators allow a genuine evaluation across continuous spatial domains; consequently leading to a stronger generalization across various mesh resolutions. An important work in this direction is FNO~\cite{li2021}, which performs a computationally efficient super-evaluation by transforming the computation onto a Fourier domain.

In this work, we present \textsc{FNOpt}, a novel self-supervised cloth simulation framework that combines meta-learning capabilities of~\cite{Nils2024} and efficient super-evaluation capabilities of FNOs. Therefore, the proposed framework enables a generalized meta-learning of various cloth motions across a diverse range of mesh resolutions. The super-evaluation capability of FNOs incorporated in cloth simulation allows a trained model on a lower resolution to perform a zero-shot generalization on a higher resolution (see \Cref{fig:teaaser}), capturing finer details that may not necessarily be present on the training data at lower resolution. Our experiments demonstrate a superior generalization and accuracy of \textsc{FNOpt} across a wide range of mesh resolutions, templates and boundary conditions compared to state-of-the-art methodologies in both supervised and self-supervised domains. 

\section{Related Work}\label{sec:Related-Work}
\noindent \textbf{Cloth Simulation.}
Early works~\cite{Terzopoulos87} modeled cloth dynamics using a continuum formulation, in which the potential energy functions were derived based on elasticity theory. The cloth was then discretized into a rectangular mesh and the resulting ODE of motion was solved by applying semi-implicit integration method.~\cite{Baraff1998} further developed an implicit integration formulation on triangular cloth and handled contact and geometric constraints directly, allowing larger step-size control while keeping the simulation stable. Although improving the efficiency and stability for real-time application, the method still fixes the cloth mesh topology, which prohibits realistic deformation with fine wrinkles and folds. To overcome this limitation,~\cite{Narain2012} presented a dynamic remeshing technique to capture high details of the cloth. It has been widely used and has become a standard framework, titled ARCSim, in cloth simulation community. 

In order to accelerate simulation, data-driven methods train neural networks in a \emph{supervised} manner using large-scale datasets containing ground truth cloth trajectories simulated by offline simulators.~\cite{Bertiche2020, Patel2020, Santesteban2019, Pan2022, Gundogdu2020, Santesteban21, Vidaurre2020, Zhang2021, Tiwari2023a} focus on garments draped over articulated bodies and regress vertex-wise displacements from a reference template, conditioned on pose and shape parameters. In these settings, the cloth behavior is largely dominated by the underlying body geometry and motion. 
To fully capture the physical interactions arising between cloth vertices themselves, Libao \cite{Ian2023} trains graph-based neural networks on trajectories generated by offline simulators. Their model builds on MeshGraphNets \cite{Pfaff21}, which has shown strong performance across various mesh-based simulation tasks. However, collecting high-quality simulation data is costly and time-consuming. Beyond the data burden, the trained networks tend to overfit to specific mesh resolutions, boundary conditions and motion patterns seen during training, which limits their ability to generalize under distribution shifts such as finer meshes or faster motions. Because explicit physical constraints are not enforced, their predictions may deviate from physically accurate behavior in unseen scenarios.

Modern advances on garment and cloth simulation perform \emph{self-supervised} learning by leveraging the key observation that the backward Euler solution to the discretized equations of motion can be recast as an optimization problem \cite{Santesteban2022,Chen2024,Stotko2024, Bertiche2022}. Hence, by training a neural network that minimizes the physics-based energy functions, we obtain a differentiable neural simulator without the need to use data simulated a priori. This relaxes the requirement on simulating large-scale dataset by physics-based simulators for training supervised models. Moreover, the per-time-step numerical integration methods used in physics-based simulators are shifted to a forward network pass. Such novel formulation is remarkable for cloth simulation, especially in the contemporary era of deep learning and large models. 
While these methods inherit the efficiency of neural networks, their predictions can be unstable due to error accumulation over long sequences. \cite{Nils2024} proposed to train a neural optimizer rather than a neural simulator, allowing for iterative updates to refine the prediction to higher precision. However, such self-supervised models are trained with a fixed rectangular mesh and can struggle when simulating cloth in significantly different mesh resolutions, limiting their usage in practical scenarios. In this paper, we overcome this limitation of self-supervised cloth simulation by incorporating resolution-agnosticism using the neural operators.

\noindent \textbf{Neural Operators for solving PDEs.}

Standard neural network architectures depend heavily on the discretization and have difficulty in generalizing to resolutions other than the training data. Recently, neural operators have been developed as a class of models that guarantee discretization invariance \cite{Li2020, li2021, kovachki2023neural}. Such property is crucial in the context of solving PDEs, where the training data is often provided at varying resolutions and high-resolution data is expensive to generate. This property is also needed for resolution-agnostic cloth simulation, allowing for simulating fine-level details such as folds and wrinkles depending on the mesh-resolution.

Various architectures of neural operators have been studied. Graph neural operators (GNO) \cite{Li2020} performed kernel integration on graph structures and can handle irregular geometries. However, as other graph-based methods, GNO is limited by computational complexity with long-range global interactions on the graph. To overcome this, FNO was introduced to represent the kernel integration in spectral domain by leveraging Fourier transform, enabling to use discrete fast Fourier transform (FFT) to improve efficiency. From these outstanding properties, FNO has become a standard in scientific computing and been applied into various domains including fluid and solid mechanics \cite{lizj2022, Choubineh2023, khorrami2025}, geoscience \cite{Yang2021, Wen2023}, weather forecasting \cite{pathak2022, Kurth2023} and inverse-design problems \cite{Zhou2024}. Their advanced computational accuracy and widespread applicability motivates their usage  in \textsc{FNOpt}.

\begin{figure*}
    
\centering
\includegraphics[width=1\linewidth]{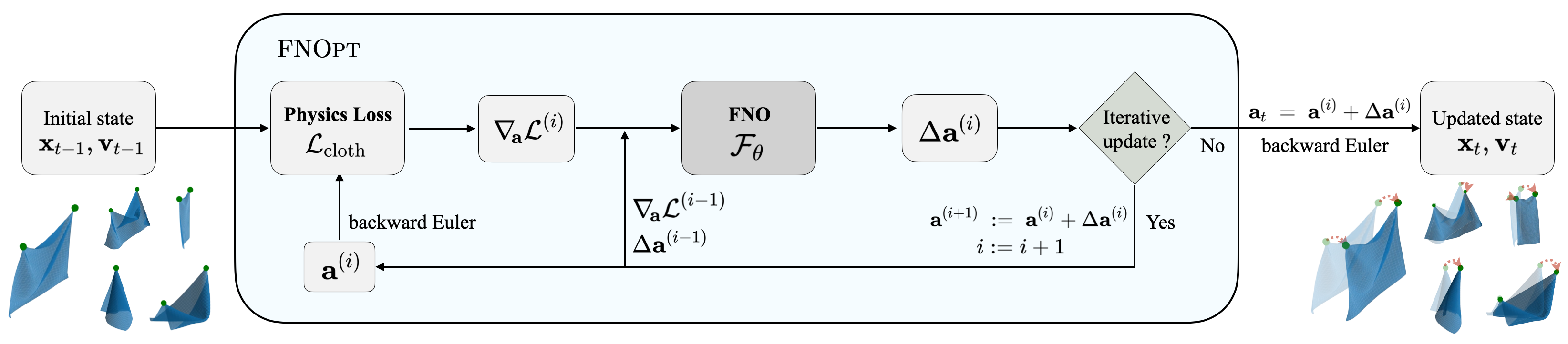}
   \caption{\textsc{FNOpt} pipeline.  At each simulation time step $t$, an inner optimization loop uses an FNO‑based optimizer to predict updates $\Delta \mathbf{a}^{(i)}$ from physics-based loss gradients and state information. After \(N\) iterations, backward Euler advances the state to $(\mathbf{x}_t, \mathbf{v}_t)$. Superscript $i$ indicates the inner iteration index.
 }
   \label{fig:pipeline}
   
\end{figure*}

\section{Simulating Cloth via \textsc{FNOpt}}\label{sec:method}
\subsection{Mathematical Foundation}\label{sec:mathematical-foundation}
Our goal is to simulate cloth dynamics, which can be modeled by the following time-varying partial differential equation (PDE) of motion
\begin{align}
    \dfrac{\partial}{\partial t}\left(\mu\dfrac{\partial x}{\partial t}\right)+\gamma\dfrac{\partial x}{\partial t} + \dfrac{\delta\mathcal{E}}{\delta x}=f(x,t),
\end{align}
where $x(p,t)$ is the position of the particle $p$ within the domain $\Omega$ at time $t$, $\mu(p)$ is the mass density, $\gamma(p)$ is the damping density, $\mathcal{E}(x)$ is the potential energy of elastic deformation and $f(x,t)$ represents external forces~\cite{Terzopoulos87}. 

By defining the state $u(p,t):=\left(x(p,t),\dfrac{\partial x(p,t)}{\partial t}\right)^\top$, we can reformulate cloth dynamics to the PDE problem
\begin{align}
    \dfrac{\partial u}{\partial t}&=\mathcal{R}\label{eq:PINO-physics}(u,t),\quad\text{ in } \Omega \times (0, \infty),\\
    u&=g,\qquad\qquad\text{in } \partial\Omega \times (0, \infty),\label{eq:PINO-ic}\\
    u&=a,\qquad\qquad\text{in } \bar{\Omega} \times \{0\},\label{eq:PINO-bc}
\end{align}
where $\mathcal{R}$ is a possibly nonlinear partial differential operator, $g$ is a known boundary condition on the boundary $\partial\Omega$ describing the handle points and their trajectories specified as input, and $a=u(\cdot, 0),$ is the initial condition describing position and velocity of  points within the closed domain, $\bar\Omega$. 
Such dynamical system can be solved by 
training a model $u_\theta$ to minimize the residual errors of \Cref{eq:PINO-physics,eq:PINO-ic,eq:PINO-bc}. In order to train it in a discretization-invariant way,
we employ an FNO, $\mathcal{F_\theta}$ \cite{li2021,li2024physics}, which maps from the initial condition $a$ to the state $u=\mathcal{F}_\theta(a)$. In contrast to standard neural networks, this choice of architecture allows us to learn a mapping $a\mapsto u$ between function spaces, which is the key to generalize to perform super-evaluation on higher-resolution cloth to get finer details.

\subsection{Training FNO}
Because the dynamics in \Cref{eq:PINO-physics,eq:PINO-bc,eq:PINO-ic} evolves over time, we must include the current time $t$ as an input to the neural operator $\mathcal{F}_\theta$ so that it can produce at that time the corresponding cloth state $u(\cdot,t)=\mathcal{F}_\theta(a,\,t)$. However, this requires to train the network within a fixed time interval $(0,T]$ for some choice of large $T$, which prevents simulation for out-of-distribution $t>T$. Therefore, as suggested by~\cite{brandstetter2022message}, we instead train an autoregressive neural operator $\mathcal{F}_\theta$ that maps current state $u(\cdot,t)$ to next state $u(\cdot,t+\Delta t)$ for initial condition $u(\cdot,0)=a$, allowing us to simulate up to any time step.

As mentioned in \Cref{sec:mathematical-foundation}, we must train the neural operator by minimizing the residual errors of \Cref{eq:PINO-physics,eq:PINO-ic,eq:PINO-bc}. Since the initial and boundary conditions associated with each sequence of cloth motion are known, the computation of the residual errors for \Cref{eq:PINO-ic,eq:PINO-bc} is straightforward. However, the physics prior in \Cref{eq:PINO-physics} requires an expensive computation of high-order derivatives of the network.  
To simplify the computation, we discretize the PDE in \Cref{eq:PINO-physics} by using  backward Euler method. Thus, we write 
\begin{align}
    \textbf{M}\dfrac{\textbf{x}_{t+1}-\textbf{x}_t-\Delta t\textbf{v}_t}{\Delta t^2}+\dfrac{\partial \mathcal{E}_\text{int}}{\partial \textbf{x}}=\textbf{f}_\text{ext}\left(\textbf{x}_{t+1},\dfrac{\textbf{x}_{t+1}-\textbf{x}_t}{\Delta t}\right),
\end{align}
where we have incorporated the damping term into external force $\textbf{f}_\text{ext}$, $\textbf{x}$ and $\textbf{v}$ are the positions and velocities, respectively. As suggested by \cite{gast2015optimization,Santesteban2022,Stotko2024,Nils2024}, solving this system can be recast as minimizing the following loss function
\begin{align}
    \mathcal{L}_\text{cloth}=\mathcal{E}_\text{int}+\mathcal{E}_\text{ext}+\mathcal{E}_\text{inertia}.\label{eq:losses}
\end{align}
The first term $\mathcal{E}_{\text{int}}(\textbf{x}_t)$ corresponds to the internal potential energies of  cloth, including stretching, shearing, and bending. The second term $\mathcal{E}_{\text{ext}}(\textbf{x}_t)$ refers to the external forces such as gravity and wind. The last energy $\mathcal{E}_{\text{inertia}}$ imposes the inertia constraint, which makes wrinkles and dynamic behavior appear. We refer to \cite{Stotko2024} for more details on the energies.


\subsection{Training Neural Optimizer via Meta-learning}\label{sec:neural-optimizer}
Using the loss functions formalized above, we  train a neural cloth simulator that predicts the acceleration $\textbf{a}_{t+1}$ from the current 
position $\textbf{x}_t$ and velocity $\textbf{v}_t$, given some boundary conditions. We use the training cycle proposed by~\cite{Stotko2024}, where the data pool is initialized with random cloth states and then progressively augmented with the model's own predictions. This technique allows us to train the model in a self-supervised manner without using any data generated from traditional physics-based simulator (PBS). However, the predictions can be unstable due to error accumulation over frames when simulating long
sequences. Hence, we follow \cite{Nils2024} to train a neural optimizer via meta-learning technique. Concretely, rather than predicting the acceleration directly, we initialize the prediction as $\textbf{a}_{t+1}^{(0)}=0$ and iteratively optimize it using the following update rule
\begin{align}
    \textbf{a}_{t+1}^{(i + 1)}=\textbf{a}_{t+1}^{(i)}+\Delta \textbf{a}_{t+1}^{(i)},\qquad\text{for }i=0,...,N-1,\label{eq:update-rule}
\end{align}
where $N$ is the number of iterations used for each time step $t + 1$, and the update direction $\Delta \textbf{a}_{t+1}^{(i)}$ is given by the neural operator $\mathcal{F}_\theta$. The resulting acceleration is then used to evaluate the next cloth state using backward Euler method as
\begin{align}
    \textbf{v}_{t+1}&=\textbf{v}_t+\Delta t\textbf{a}_{t+1},\\
    \textbf{x}_{t+1}&=\textbf{x}_t+\Delta t\textbf{v}_{t+1}.
\end{align}

Combining all features introduced so far, our \textsc{FNOpt} framework follows the pipeline in \Cref{fig:pipeline}.

\subsection{Inference}
At inference time, given initial position and velocity of the cloth as well as boundary conditions including handle points and their trajectories, \textsc{FNOpt} performs an autoregressive rollout. For each time step, we apply the learned optimizer $\mathcal{F}_{\theta^\star}$ for a fixed number $N$ of inner updates to minimize the objective in \Cref{eq:losses}.
Note that the same trained $\mathcal{F}_{\theta^\star}$ is used across all experiments, no fine-tuning is required. 

Optionally, to handle self-collision between cloth, we can augment the objective with a repulsive term, $\mathcal{L_\text{rep}}$ \cite{Lee2023} at inference time, which reduces interpenetration without retraining: 


\[
\mathcal{L}_{\text{rep}}
= \lambda_{\text{rep}} \sum_{i=1}^{N} \sum_{j \in \mathcal{A}_i}
  -\log\!\bigl(\|\mathbf{v}_i - \mathbf{v}_j\|^{2}\bigr).
\]
Where $\mathcal{A}_i \coloneqq \{\, j \nsim i : \|\mathbf{v}_i-\mathbf{v}_j\|<\delta \,\}$. Here \(j \nsim i\) denotes non-adjacent vertices \((i,j)\notin E\) for edge set \(E\).
We set \(\delta\) to the local grid spacing; see \Cref{sec:supp_ablation_repulsive} for ablations.

\section{Experiments}\label{sec:Experiments}

We use the official implementation provided in the NeuralOperator library \cite{kossaifi2024} which includes the FNO. 
Regarding the network hyper-parameters, we use $4$ Fourier layers, with the number of Fourier modes set to $8$ for both spatial dimensions (height and width). The number of hidden channels for intermediate Fourier layers, the lifting layer, the projection layer are set to $64$, $256$, $64$, respectively. We train only at $32\times32$ resolution, with data pool of $1000$ data points updated progressively during the self-supervised training. We set the batch size to $10$ and learning rate to $10^{-3}$ and train the network for $100$ epochs, which takes $8$ hours on an NVIDIA H100 GPU.
For experiments, we use the dataset from \cite{Ian2023} which contains trajectories of a square cloth generated by ARCSim given sequences of handle point trajectory, and run the evaluation on NVIDIA A100 GPU. We use the evaluation set which contains 11 trajectories for translation and $4$ trajectories for rotation run at $60$ fps, which serves as ground truth. We selected various baselines for comparison: 1) the self-supervised cloth model used in PGSfT \cite{Stotko2024}; 2) Metamizer \cite{Nils2024}, the neural optimizer baseline with iterative refinement; and 3) MeshGraphNetRP  (MGNRP) \cite{Ian2023}, the supervised baseline trained using PBS data. We use the official checkpoints released by the respective authors, except for Metamizer, which is re-trained exclusively with the cloth data pool to ensure a fair comparison.
The physics-based losses in \Cref{eq:losses} are different from ARCSim, which leads to different meaning of material coefficients in our setting. Hence, we apply grid search to find the corresponding material for PGSfT, Metamizer and \textsc{FNOpt} on the simulated sequences. Unless otherwise noted, the repulsive term is disabled. We refer to \Cref{sec:supp_ablation_repulsive} for experiments with the repulsive loss.

\subsection{SOTA Comparison}\label{sec:sota-comparison}

\paragraph{Training resolution.}
We first evaluate \textsc{FNOpt} at the training resolution of $32\times 32$. We measure the chamfer distance ($e_\text{CD}$) between the ground truth and the predicted point clouds, each of which is uniformly sampled $10^4$ points from the corresponding mesh. As shown in \Cref{tab:res_chamfer}, compared to other self-supervised methods, \textsc{FNOpt} significantly outperforms PGSfT and surpasses Metamizer in most of the sequences. 
MGNRP is trained with supervision on domain-matched ground truth trajectories at the same resolution and therefore attains the lowest $e_{\text{CD}}$ in-domain.
However, as we will show in later sections, this advantage does not transfer to out-of-domain settings, making it less scalable.

\begin{figure}[ht!]
  \centering
\includegraphics[width=1\linewidth]{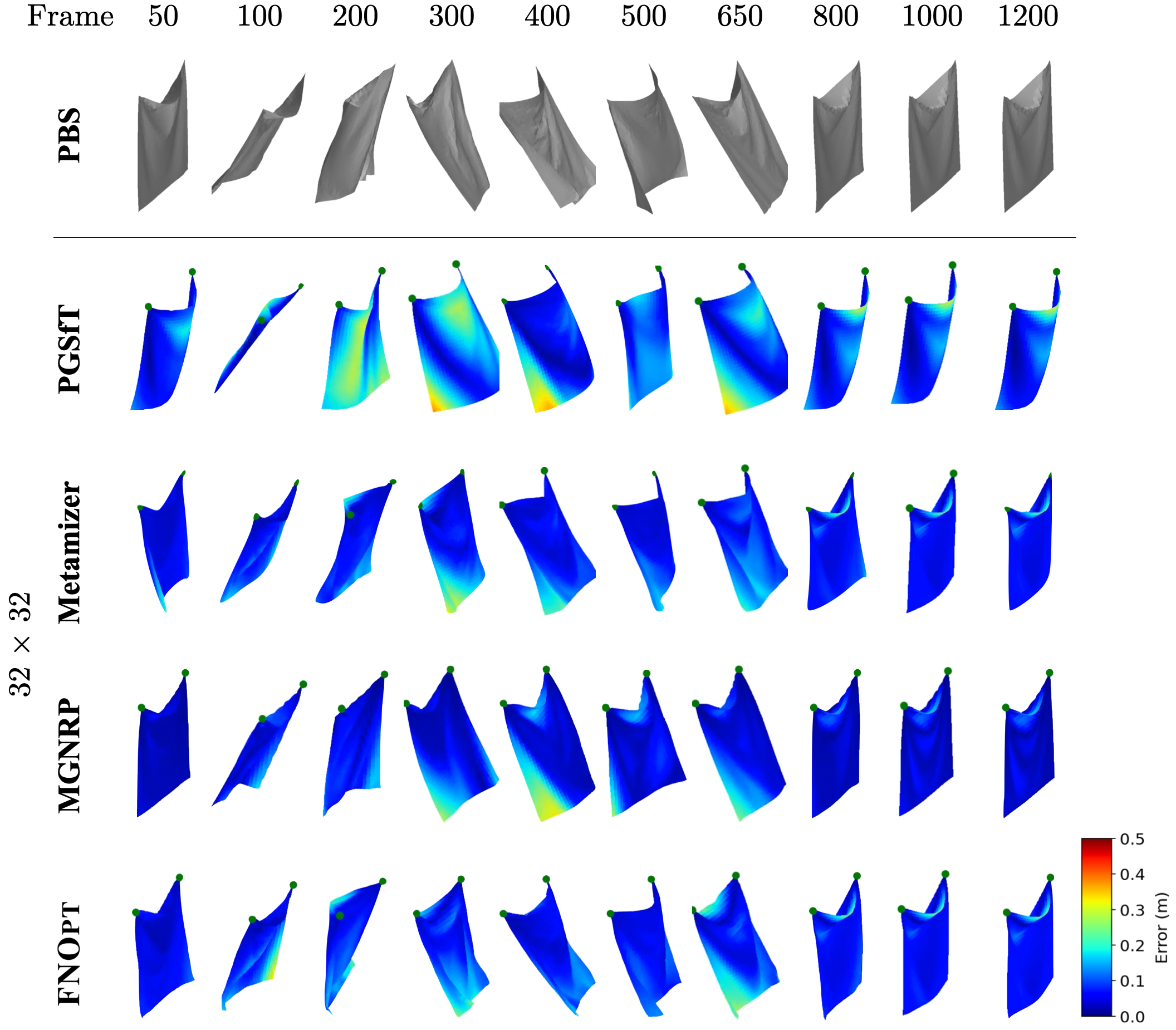}
   \caption{Vertex-wise error maps on the \texttt{xy\_v2} sequence on training resolution. Colors indicate the Euclidean distance between rollout and PBS. }
   \label{fig:error_map}
\end{figure}

\Cref{fig:error_map} shows the evaluated shapes and their Euclidean distance to the PBS ground truth sequence \texttt{xy\_v2}. PGSfT suffers from noticeable drift over long horizons 
resulting in high global errors. In addition, it fails to recover high-frequency wrinkles, leading to unnaturally stiff cloth behavior. Metamizer and \textsc{FNOpt} perform quite well at training, showing good dynamics and wrinkles. Although achieving the best quantitative results, MGNRP exhibits subtle but noticeable temporal jitters (visible in supplementary video). \Cref{fig:framewise_metrics} (top) plots the per-frame chamfer distance $e_{\text{CD}}$ on \texttt{xy\_v2}; PGSfT produces significantly higher error than the other methods, showing limited dynamic fidelity and a lack of fine-scale detail.

\begin{figure}[ht!]
  \centering
\includegraphics[width=1\linewidth]{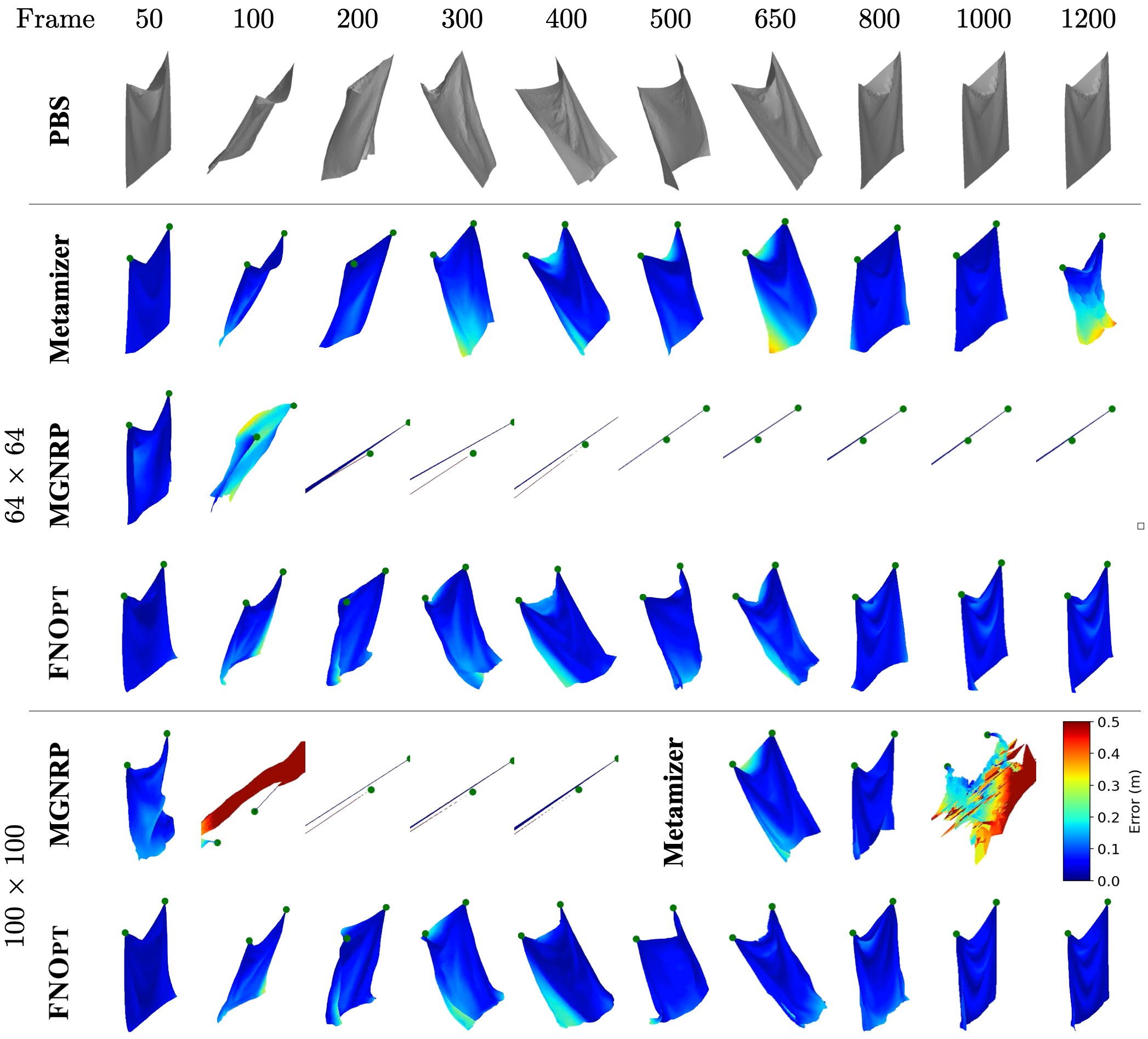}
   \caption{Vertex-wise error maps on finer resolutions on the \texttt{xy\_v2} sequence.}
   \label{fig:error_map_finer_resolution}
\end{figure}

\paragraph{Super-evaluation.}
To show our generalizability in multi-resolution, we perform simulation with higher resolutions $64\times64$ and $100\times100$. Although using graph-based network that can deal with different mesh topologies, MGNRP fails to simulate at higher resolutions. PGSfT's U-Net implementation supports inference only at the training resolution, cross-resolution results are unavailable. Metamizer also uses U-Net but their implementation can be run on different resolutions. As shown in \Cref{tab:res_chamfer}, Metamizer is still able to get decent results at $64\times 64$ but struggles at $100\times 100$. Thanks to its FNO architecture, \textsc{FNOpt} can perform super-evaluation automatically at both $64\times 64$ and $100\times 100$. Since the wrinkles and details are simulated more precisely in these cases, it improves the average chamfer distance to $4.8$ and $4.7$ respectively, bridging the gap with respect to supervised method MGNRP.
 We show comparison for super-evaluation of sequence \texttt{xy\_v2} in \Cref{fig:error_map_finer_resolution}.
 Metamizer exhibits large errors and temporal instabilities on a finer $64\times64$ mesh, and diverges at $100\times100$. MGNRP also results in divergence at both $64\times 64$ and $100\times 100$ resolutions. In contrast, \textsc{FNOpt} produces visually plausible cloth with realistic wrinkles and maintains stable, accurate rollouts across all tested resolutions, demonstrating superior fidelity and resolution-agnostic generalization (more details in supplementary video). Furthermore, the per-frame chamfer distance in the lower figure of \Cref{fig:framewise_metrics} demonstrates our stability when simulating higher resolutions.
 We also computed the relative 3D error between the ground truth ($e_\text{3D}$) and predicted meshes;
more details can be found in \Cref{sec:supp_sota_comparison} in supplementary.

\paragraph{Interpretation of the results.}
We analyze why the different approaches attain varying levels of motion realism, fine-scale detail, and temporal stability.
PGSfT~\cite{Stotko2024} is trained in a fully self-supervised fashion. Its one-shot architecture, however, cannot correct accumulated errors, leading to noticeable drift and a loss of high-frequency wrinkles in long rollouts.
MGNRP~\cite{Ian2023} is trained in a supervised fashion on pre-computed ground truth trajectories, achieving high accuracy for motions, velocities, and mesh resolutions seen during training. It can handle irregular meshes due to its use of the MeshGraphNets architecture. However, the model extrapolates poorly, and its performance degrades significantly when tested on finer resolutions or faster motions outside the training distribution.
Metamizer~\cite{Nils2024} improves visual fidelity and wrinkle detail over PGSfT by employing an iterative, meta-learned optimizer. Nevertheless, its performance drops sharply on finer meshes, and rollouts exhibit temporal instabilities.
\textsc{FNOpt} produces stable, high-fidelity rollouts across all tested resolutions. We attribute this robustness to the resolution-agnostic FNO backbone, which supports zero-shot inference on previously unseen mesh resolutions, and to the meta-learned optimizer that optimizes per-step acceleration with only a few iterations.

\paragraph{Runtime performance.}
All timings were measured on an NVIDIA A100 GPU with a 1024-vertex mesh. With the default setting of $10$ optimizer iterations, \textsc{FNOpt} runs at $61~\mathrm{ms}$/frame, and at $33~\mathrm{ms}$/frame with $5$ iterations at the expense of a modest loss in accuracy. Enabling the repulsive term incurs an additional $3~\mathrm{ms}$/frame. In comparison, the supervised baseline MGNRP completes a frame in $40~\mathrm{ms}$ using the authors’ released model, while PGSfT, which requires only a single feed‑forward pass, achieves $7~\mathrm{ms}$/frame but fails to capture fine‑scale wrinkles. Metamizer with iterative refinement runs at $63~\mathrm{ms}$/frame.
The offline finite-element solver ARCSim requires $404~\mathrm{ms}$/frame. Overall, \textsc{FNOpt} offers a compelling trade‑off: its per-frame runtime is about $1.5\times$ that of the supervised baseline while delivering substantially higher accuracy on cross-resolution rollouts, and is roughly $6\times$ faster than the PBS.

\begin{table*}[t]
\centering
\setlength{\tabcolsep}{4pt}
\begin{tabular}{l|ccc| c| ccc| ccc}
\toprule
\multirow{3}{*}{\makecell{Sequence\\name}} 
& \multicolumn{3}{c}{MGNRP \cite{Ian2023}} 
& \multicolumn{1}{c}{PGSfT \cite{Stotko2024}} 
& \multicolumn{3}{c}{Metamizer \cite{Nils2024}} 
& \multicolumn{3}{c}{\textsc{FNOpt}[Ours]} \\
\cmidrule(lr){2-4}\cmidrule(lr){5-5}\cmidrule(lr){6-8}\cmidrule(lr){9-11}
& \multicolumn{1}{c}{res32} & \multicolumn{1}{c}{res64} & \multicolumn{1}{c}{res100}
& \multicolumn{1}{c}{res32}
& \multicolumn{1}{c}{res32} & \multicolumn{1}{c}{res64} & \multicolumn{1}{c}{res100}
& \multicolumn{1}{c}{res32} & \multicolumn{1}{c}{res64} & \multicolumn{1}{c}{res100} \\
\midrule
\texttt{xy\_v2}       & \textbf{4.8} & \cc & \cc & 25.3 & \ul{7.1}  & \ul{33.6} & \cc & \ul{7.1}  & \textbf{5.2} & \textbf{4.2} \\
\texttt{xy\_v2\_opp}  & \textbf{5.0} & \cc & \cc & 25.2 & 7.7       & \ul{14.7} & 12.9& \ul{6.5}  & \textbf{5.3} & \textbf{4.8} \\
\texttt{yz\_v2}       & \textbf{2.6} & \cc & \cc & 10.7 & 11.6      & \ul{3.6}  & 77.2& \ul{4.9}  & \textbf{3.4} & \textbf{5.5} \\
\texttt{yz\_v2\_opp}  & \textbf{3.0} & \cc & \cc & 13.5 & 4.9       & \ul{4.2}  & \cc & \ul{4.3}  & \textbf{3.2} & \textbf{3.6} \\
\texttt{xz\_v2}       & \textbf{2.7} & \cc & \cc & 22.5 & \ul{6.5}  & \textbf{6.6}  & \cc & 8.9       & \textbf{6.6} & \textbf{3.9} \\
\texttt{xyz\_v2}      & \textbf{5.4} & \cc & \cc & 21.9 & \ul{6.3}  & \textbf{4.9}  &29.9& 6.4       & \textbf{4.9} & \textbf{4.3} \\
\texttt{xyz\_v2\_opp} & \textbf{4.3} & \cc & \cc & 22.7 & 6.4       & \ul{4.9}  & \cc & \ul{6.4}  & \textbf{4.1} & \textbf{5.2} \\
\texttt{xyz\_v3}      & \textbf{3.5} & \cc & \cc & 21.4 & 7.2       & \ul{10.2} & \cc & \ul{6.4}  & \textbf{4.7} & \textbf{3.7} \\
\texttt{xyz\_v3\_opp} & \textbf{3.0} & \cc & \cc & 20.6 &29.2       & \ul{12.1} & \cc & \ul{6.1}  & \textbf{4.8} & \textbf{4.6} \\
\texttt{xyz\_v4}      & \textbf{3.7} & \cc & \cc & 18.9 & 7.2       & \ul{4.8}  &21.2& \ul{6.7}  & \textbf{3.7} & \textbf{3.5} \\
\texttt{xyz\_v4\_opp} & \textbf{3.3} & \cc & \cc & 19.0 & \ul{5.8}  & \ul{6.2}  & \cc & 7.4       & \textbf{4.2} & \textbf{4.7} \\
\texttt{rot\_h0}      & \textbf{5.4} & \cc & \cc & 17.5 & 8.2       & \ul{21.1} & \cc & \ul{6.9}  & \textbf{5.2} & \textbf{6.2} \\
\texttt{rot\_h0\_opp} & \textbf{5.7} & \cc & \cc & 17.5 & 9.3       & \ul{6.8}  & \cc & \ul{7.6}  & \textbf{6.6} & \textbf{5.9} \\
\texttt{rot\_h1}      & \textbf{5.1} & 8.3 & \cc & 18.6 & 7.8       & \ul{7.6}  & \cc & \ul{7.6}  & \textbf{4.5} & \textbf{6.0} \\
\texttt{rot\_h1\_opp} & \textbf{5.3} & 9.4 & \cc & 16.7 & 8.3       & \ul{7.7}  & \cc & \ul{7.5}  & \textbf{5.7} & \textbf{6.2} \\
\midrule
\rowcolor[gray]{0.92}
Avg ± $\sigma$         &     \avgstd{\textbf{4.2}}{1.1} & \cc & \cc & \avgstd{19.5}{3.9} & \avgstd{8.9}{5.6}      & \avgstd{\ul{10.0}}{7.8 } & \cc & \avgstd{\ul{6.7}}{1.1}  & \avgstd{\textbf{4.8}}{1.0} & \avgstd{\textbf{4.7}}{0.9} \\
\bottomrule
\end{tabular}
\caption{Comparison of the chamfer distance $e_\text{CD}$ (scaled by $10^{3}$) at different resolutions for each motion sequence. The best and second-best results are shown in \textbf{bold} and \underline{underline}, respectively.}
\label{tab:res_chamfer}
\end{table*}

\begin{figure}[ht!]
  \centering
\includegraphics[width=1\linewidth]{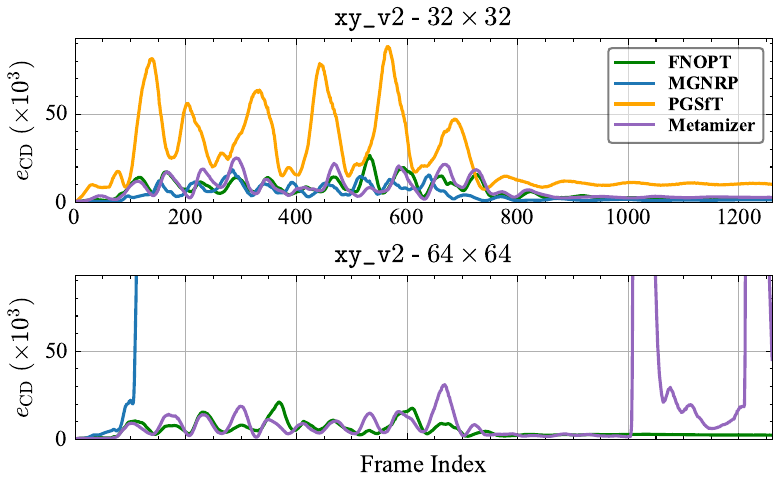}
   \caption{Per-frame $e_\text{CD}$ ($\times 10^{\!3}$) on \texttt{xy\_v2} sequence. All models are trained at $32\times 32$. \textsc{FNOpt} achieves second lowest error on $32\times32$ resolution and generalizes to the finer $64\times64$ resolution.}
   \label{fig:framewise_metrics}
   
\end{figure}

\subsection{Boundary Condition Generalization} \label{sec:bc-generalization}
Beyond scaling across resolutions, \textsc{FNOpt} remains robust under changes to boundary conditions (e.g. handle speed and placement, mesh sizes, etc), in contrast to baseline methods that struggle in such settings. 

\begin{figure}[ht!]
  \centering
\includegraphics[width=1\linewidth]{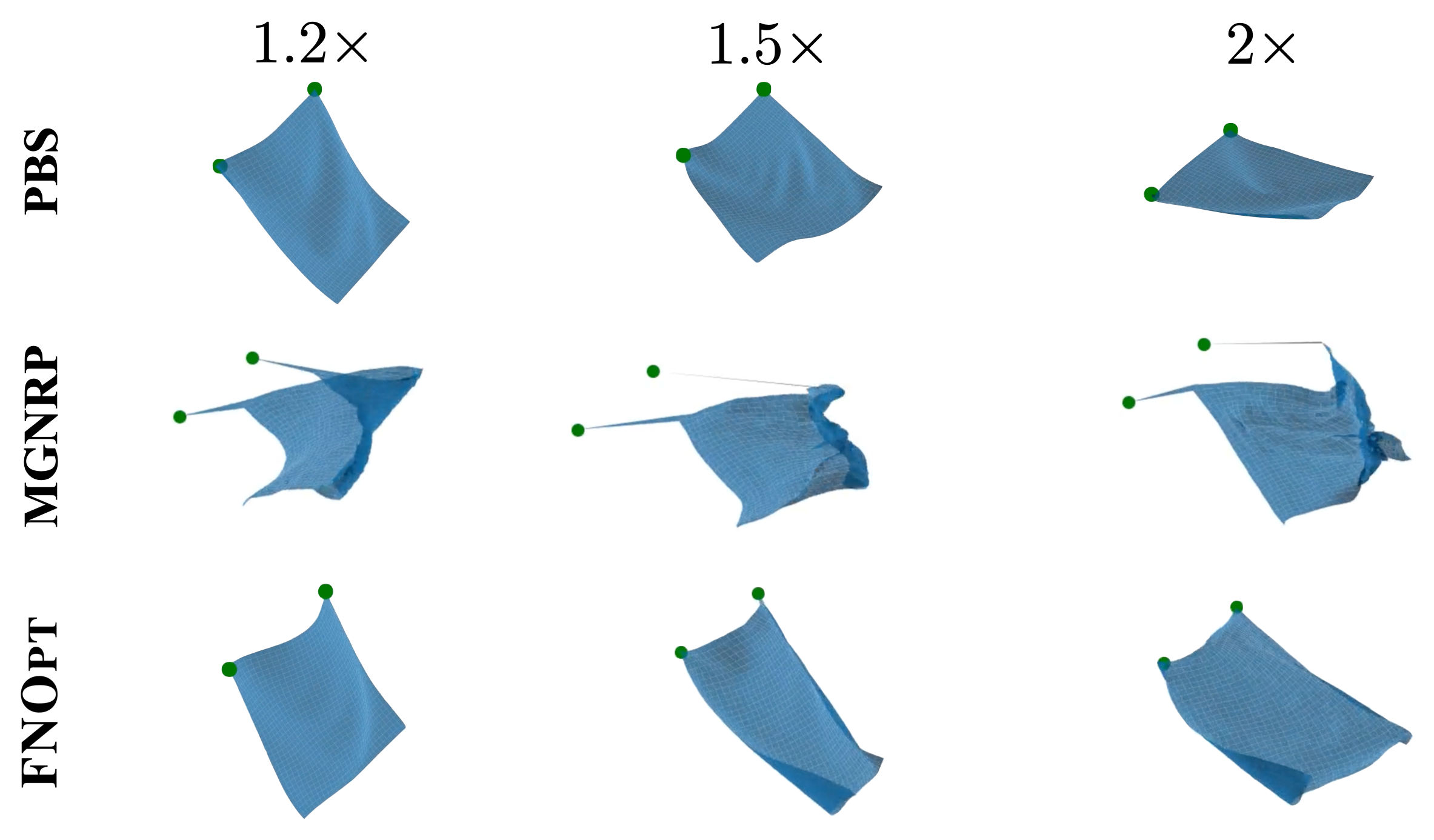}
\caption{Generalization to speeds on a representative frame of the \texttt{xy\_v2} rollout. MGNRP exhibits overstretching artifacts near the handles while \textsc{FNOpt} remains stable. }
   \label{fig:speed}
\end{figure}

\paragraph{Speed generalization.} We evaluate the speed generalization ability of \textsc{FNOpt} and compare with the supervised approach MGNRP. We create new motion sequences by accelerating the original ones using interpolation. We test speed factors $\alpha \in \{1.2, 1.5, 2\}$. The results are shown in \Cref{fig:speed}. Both methods are able to handle slight speed increase, as shown in $1.2\times$ with no noticeable artifacts. When speed factor goes to $1.5\times$, MGNRP shows overstretching artifacts around upper corners. With $2\times$ speed, severe artifacts are shown and MGNRP simulation becomes unstable. These behaviors happen because the accelerated sequences are not present in the training set, preventing MGNRP from generalizing to out-of-distribution motions. On the other hand, \textsc{FNOpt} remains stable at all tested motion speeds without artifacts around handles. 
 
\paragraph{Flexible handle placement.}
We evaluate 5 different configurations of the handle points: \emph{Diagonal}, \emph{Mid-Edge}, \emph{Corner-Center}, \emph{Single Mid-Edge} and \emph{Single Center}. \emph{Diagonal} puts the two handles on the opposite corners. For \emph{Mid-Edge}, we fix the handles on two midpoints of the opposite sides of the mesh. \emph{Corner-Center} uses one corner and the center of the cloth as handles. Each of the two settings \emph{Single Mid-Edge} and \emph{Single Center} uses only a single handle point on the midpoint of the upper edge and the center of the cloth, respectively. 

\begin{figure}[htbp]
  \centering
\includegraphics[width=1\linewidth]{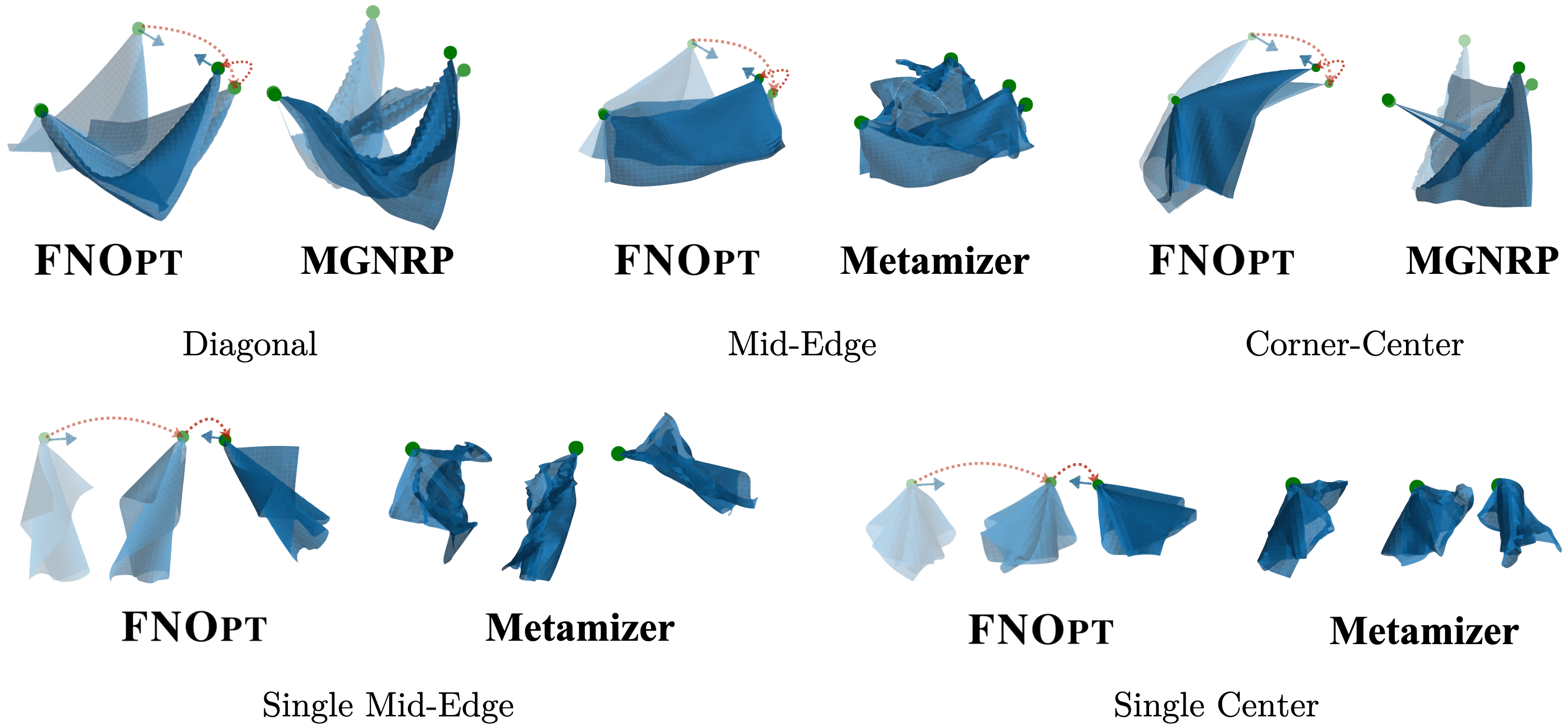}
\caption{
Qualitative comparison of generalization to various handle configurations.
The top row shows configurations with two control points. The bottom row contains single-handle scenarios.
\textsc{FNOpt} consistently produces plausible and stable cloth dynamics across all settings. Green points denote control handles. Red dashed arrows denote handle trajectories. Blue arrows indicate acceleration directions. Repulsive loss is activated.
}
\label{fig:handle_points}
\end{figure}

The qualitative comparison is illustrated in \Cref{fig:handle_points}. We can see that \textsc{FNOpt} supports arbitrary handle placements as boundary conditions, yielding stable rollouts with preserved fine‑scale wrinkles without retraining or further fine-tuning. MGNRP only supports two-handle configuration with both handles at the top corners, and fails to simulate these 5 configurations. Metamizer utilizes the same formulation of self-supervised neural optimizer as \textsc{FNOpt}, but yields unstable simulation. 
In addition, \textsc{FNOpt} is generalizable to non-square cloth meshes, see \Cref{sec:supp_boundary_condition} in supplementary.

\subsection{Ablation Studies} 

\paragraph{Iterative Update.}
The iterative update discussed in \Cref{sec:neural-optimizer} is crucial for our framework. The improvement is already demonstrated in Metamizer results (see \Cref{tab:res_chamfer}, \Cref{fig:error_map,fig:framewise_metrics}), in which the iterative update is employed to achieve higher precision compared to PGSfT. We further train a baseline model using PGSfT formulation that directly predicts the acceleration $\textbf{a}_{t+1}$ in a single forward pass, without using iterative update in \Cref{eq:update-rule}. This baseline is trained in a supervised fashion using the same loss in \Cref{eq:losses}, and uses FNO backbone rather than U-Net. As shown in \Cref{tab:ablation_chamfer}, this approach yields significantly less accurate rollouts, and fails to capture high-frequency details such as wrinkles.

\paragraph{Neural Optimizer.}
To evaluate the neural optimizer formulation in \textsc{FNOpt} framework, we replace the trainable neural optimizer by a classical optimizer using a step size and direction computed from the gradients with respect to the losses in \Cref{eq:losses}. 
This setup still performs iterative optimization, but lacks any data‑driven adaptation of step length or search direction. We evaluate two classical optimizers, Adam \cite{Kingma2014} and vanilla gradient descent (GD) \cite{Cauchy1847}, each with a learning rate sweep, and report their best-performing results in \Cref{tab:ablation_chamfer}. Even with a tuned step size, GD fails to minimize $\mathcal{L}_{\text{cloth}}$ effectively and yields poor rollouts. Adam converges given large enough iterations ($\sim$$500$), but remains $30\times$ slower (see \Cref{fig:abl_optimizer_runtime_comparison}) than ours and less accurate in $e_\text{CD}$. This shows the necessity of an adaptive, learned optimization scheme.
Additional detailed comparison, including runtime, error metrics and losses versus the number of iterations per time step, are provided in \Cref{sec:supp_ablation} in supplementary.

\paragraph{FNO Architecture.}
As discussed in \Cref{sec:mathematical-foundation}, we adopt FNO as the backbone in our framework because its global spectral kernels capture long‑range interactions and naturally support resolution‑agnostic inference, producing realistic wrinkles and faithful high‑resolution details. We have compared FNO with the U-Net backbone used by Metamizer in \Cref{sec:sota-comparison} and shown a clear advantage in both accuracy and visual quality over various resolution. We further show the benefit of using FNO over MeshGraphNets\cite{Pfaff21}, a popular graph‑based simulator in cloth and fluid dynamics. Note that we keep the same formulation as the \textsc{FNOpt} framework and only change the network architecture. We refer to this variant as MGN. Despite using identical loss objectives, iteration counts, and training strategy, the MGN version exhibits visible high‑frequency oscillations even at the training resolution and diverges rapidly on $64 \times 64$ meshes, see \Cref{fig:abl_mgn_error_map} in supplementary.

We hypothesize that the instability arises from the autoregressive rollouts: with a fixed number of message‑passing steps, MeshGraphNets can propagate information only within a limited neighborhood, so local errors are re-fed into the next step, propagate and amplify over time. Since the learned optimizer relies on accurate per‑vertex gradients, even small per‑step misestimations may accumulate and drive the system into unstable regimes. By contrast, the global spectral kernels of FNOs couple every vertex to the entire cloth surface, yielding smoother, more consistent updates and preventing long‑horizon error growth.

\begin{figure}[htb]
  \centering
\includegraphics[width=1\linewidth]{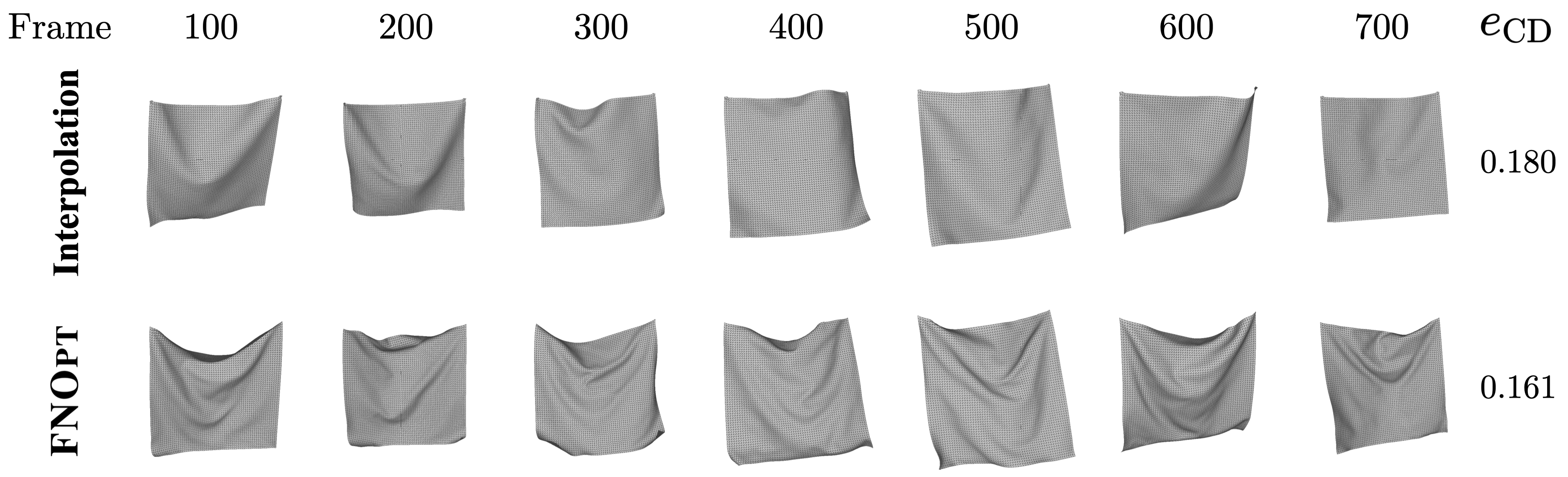}
   \caption{Visual comparison of  $64 \times 64$ rollouts using bilinear interpolation (upper row) and ours (lower row).}
   \label{fig:abl_interp}
   
\end{figure}

\paragraph{Super-evaluation.}
We emphasize that \textsc{FNOpt} framework allows for super-evaluation that is trained on a lower resolution and generalizes directly to higher resolution, capturing finer details that were not presented in the low-resolution training data. This property is essential for physics simulation, especially when simulating cloth with folds and wrinkles. We clarify that such property is different from interpolation, which estimates high-resolution rollouts from lower-resolution prediction. To evaluate, we estimate the $64 \times 64$ rollouts by bilinearly interpolating the predicted $32 \times 32$ accelerations given by \textsc{FNOpt}. As can be seen in \Cref{fig:abl_interp}, the resulting interpolation recovers fewer fine-scale wrinkles due to information loss during upsampling, showcasing the necessity of \textit{super-evaluation} at the target resolution.

\begin{table}[t]
\centering
\setlength{\tabcolsep}{8pt}
\begin{tabular}{lr}
\toprule
\textbf{Method} & $e_\text{CD}$ ↓ \\
\midrule
\textsc{FNOpt} (Ours)        & \avgstd{4.8}{0.99} \\
No iterative update        & \avgstd{24.7}{3.22} \\
GD (lr=0.01, 500 iters)      & \avgstd{419.9}{114.4} \\
Adam (lr=0.1, 500 iters)     & \avgstd{7.7}{2.6} \\
Interpolation to hi-res      & \avgstd{7.0}{1.27} \\
\bottomrule
\end{tabular}
\caption{Ablation study at $64\times64$ resolution. We report $e_{\text{CD}}$ over all evaluation sequences. Chamfer distances are multiplied by $10^3$ for readability.}
\label{tab:ablation_chamfer}
\end{table}

\paragraph{Number of iterations per time step.}
To analyze the trade-off between iteration count and performance, we evaluate our method using different value of $N$ (iterations per time step) ranging from $3$ to $30$. \Cref{fig:abl_iteration_vs_runtime} in supplementary plots the mean accuracy and runtime per frame curves at three resolutions across the tested values of $N$.
We first observe that the runtime is quasi-linear with respect to the number of iterations. Meanwhile, increasing the number of iterations from $5$ to $10$ yields a significant accuracy gain. However, increasing to 20 iterations offers no further improvement while doubling the computational cost.
The same trend is reflected in the $e_{\text{3D}}$ error. See \Cref{tab:abl_n_iteration_time} in supplementary for the quantitative results of runtimes and Chamfer distances at each iteration count and resolution. 

\Cref{tab:abl_n_iteration} reports the per-sequence accuracy for $5$, $10$, and $20$ iterations per time step across three mesh resolutions.
These results suggest that approximately $10$ iterations are sufficient for the learned optimizer to converge across all tested mesh resolutions. We therefore adopt $N=10$ as the default to balance accuracy and efficiency. Runtimes are similar across resolutions, indicating that \textsc{FNOpt} maintains efficient inference across resolutions.

\section{Conclusion}
We presented \textsc{FNOpt}, a novel resolution-agnostic cloth simulation framework that learns an optimizer from FNOs to simulate cloth dynamics. Extensive experiments demonstrate that  \textsc{FNOpt} achieves stable, accurate and efficient cloth dynamics in comparison with previous cloth simulators. Moreover, by leveraging the discretization-invariant capabilities of neural operators, it exhibits significant generalizability across a wide range of mesh resolutions and boundary conditions.

\paragraph{Limitations and future work.} 

While \textsc{FNOpt} scales across mesh resolutions, sizes and handle configurations,
the current implementation is designed for rectangular grids. It is therefore not intended to operate directly on irregular or unstructured meshes. Extending the framework to handle arbitrary mesh topology using frameworks such as GINO \cite{li2024} is left for future work.

{
    \small
    \bibliographystyle{ieeenat_fullname}
    \bibliography{main}
}

\input{supplementary}
\clearpage
\end{document}

%% file: supplementary.tex
    \clearpage
\section{Supplementary}

\subsection{SOTA Comparison} \label{sec:supp_sota_comparison}
In the main paper, we reported only the Chamfer Distance ($e_{\text{CD}}$) metric due to space limitations. Here, we complement those results by reporting the per-sequence 3D error $e_{\mathrm{3D}}$ (defined in \cite{Sidhu2020}) in \Cref{tab:res_e3d},  computed using the known vertex correspondences after re-meshing the ground truth.
Both metrics show broadly consistent trends across methods, though some variations in relative rankings can be observed.

\begin{table*}[t]
\centering
\setlength{\tabcolsep}{4pt}
\begin{tabular}{l|ccc c ccc ccc}
\toprule
\multirow{3}{*}{\makecell{Sequence\\name}} 
& \multicolumn{3}{c}{MGNRP \cite{Ian2023}} 
& \multicolumn{1}{c}{PGSfT \cite{Stotko2024}} 
& \multicolumn{3}{c}{Metamizer \cite{Nils2024}} 
& \multicolumn{3}{c}{\textsc{FNOpt}[Ours]} \\
\cmidrule(lr){2-4}\cmidrule(lr){5-5}\cmidrule(lr){6-8}\cmidrule(lr){9-11}
& \multicolumn{1}{c}{res32} & \multicolumn{1}{c}{res64} & \multicolumn{1}{c}{res100}
& \multicolumn{1}{c}{res32}
& \multicolumn{1}{c}{res32} & \multicolumn{1}{c}{res64} & \multicolumn{1}{c}{res100}
& \multicolumn{1}{c}{res32} & \multicolumn{1}{c}{res64} & \multicolumn{1}{c}{res100} \\
\midrule
\texttt{xy\_v2}       & \textbf{11.0} & \cc & \cc & 24.1 & 15.5     & \ul{27.8} & \cc & \ul{15.4} & \textbf{12.7} & \textbf{11.0} \\
\texttt{xy\_v2\_opp}  & \textbf{11.8} & \cc & \cc & 23.2 & 16.2     & \ul{17.9} & 16.5 & \ul{14.5} & \textbf{12.0} & \textbf{10.5} \\
\texttt{yz\_v2}       & \textbf{ 9.7} & \cc & \cc & 20.1 & 17.6     & \ul{12.1} & 34.3 & \ul{15.0} & \textbf{11.3} & \textbf{13.1} \\
\texttt{yz\_v2\_opp}  & \textbf{ 8.6} & \cc & \cc & 21.4 & 13.8     & \ul{12.8} & \cc & \ul{13.5} & \textbf{10.3} & \textbf{10.9} \\
\texttt{xz\_v2}       & \textbf{ 7.9} & \cc & \cc & 21.8 & \ul{13.2}& \textbf{11.9} & \cc & 15.2    & \ul{13.3} & \textbf{9.9} \\
\texttt{xyz\_v2}      & \textbf{10.7} & \cc & \cc & 21.7 & 14.7     & \textbf{11.4} & 18.9 & \ul{14.6} & \ul{12.0} & \textbf{10.5} \\
\texttt{xyz\_v2\_opp} & \textbf{10.6} & \cc & \cc & 22.7 & 15.4     & \ul{11.5} & \cc & \ul{14.7} & \textbf{10.8} & \textbf{11.4} \\
\texttt{xyz\_v3}      & \textbf{ 9.1} & \cc & \cc & 22.7 & \ul{14.5} & \ul{15.3} & \cc & 15.2    & \textbf{12.2} & \textbf{10.6} \\
\texttt{xyz\_v3\_opp} & \textbf{ 9.7} & \cc & \cc & 22.1 & 23.5 & \ul{20.7} & \cc & \ul{14.3} & \textbf{11.8} & \textbf{10.9} \\
\texttt{xyz\_v4}      & \textbf{ 9.4} & \cc & \cc & 21.9 & 14.8 & \ul{11.4} & 22.1 & \ul{14.4} & \textbf{10.7} & \textbf{9.9} \\
\texttt{xyz\_v4\_opp} & \textbf{ 9.0} & \cc & \cc & 21.4 & \ul{13.4} & \ul{13.3} & \cc & 15.2    & \textbf{11.2} & \textbf{11.2} \\
\texttt{rot\_h0}      & \textbf{12.5} & \cc & \cc & 23.5 & 17.0 & \ul{22.4} & \cc & \ul{16.1} & \textbf{14.3} & \textbf{15.1} \\
\texttt{rot\_h0\_opp} & \textbf{12.8} & \cc & \cc & 22.9 & 18.3 & \textbf{14.8} & \cc & \ul{17.1} & \ul{15.8} & \textbf{14.8} \\
\texttt{rot\_h1}      & \textbf{11.4} & 15.4 & \cc& 23.8 & 16.8 & \ul{14.9} & \cc & \ul{16.8} & \textbf{13.9} & \textbf{14.6} \\
\texttt{rot\_h1\_opp} & \textbf{11.9} & 16.4 & \cc& 23.2 & 17.8 & \textbf{15.0} & \cc & \ul{17.0} & \ul{15.8} & \textbf{14.7} \\
\midrule

\rowcolor[gray]{0.92}
Avg ± $\sigma$         & \avgstd{\textbf{10.4}}{1.4} & \cc & \cc & \avgstd{22.4}{1.0} & \avgstd{16.2}{2.5} & \avgstd{\ul{15.5}}{4.6} & \cc & \avgstd{\ul{15.3}}{1.0} & \avgstd{\textbf{12.6}}{1.7} & \avgstd{\textbf{11.9}}{1.9} \\

\bottomrule
\end{tabular}
\caption{Comparison of the 3D error $e_{\text{3D}}$ ($\times 10^{2}$; lower is better) at different resolutions for each motion sequence. Within each resolution group, the best and second-best results are shown in \textbf{bold} and \underline{underline}, respectively.}
\label{tab:res_e3d}
\end{table*}

\begin{figure}[H]
  \centering
\includegraphics[width=1\linewidth]{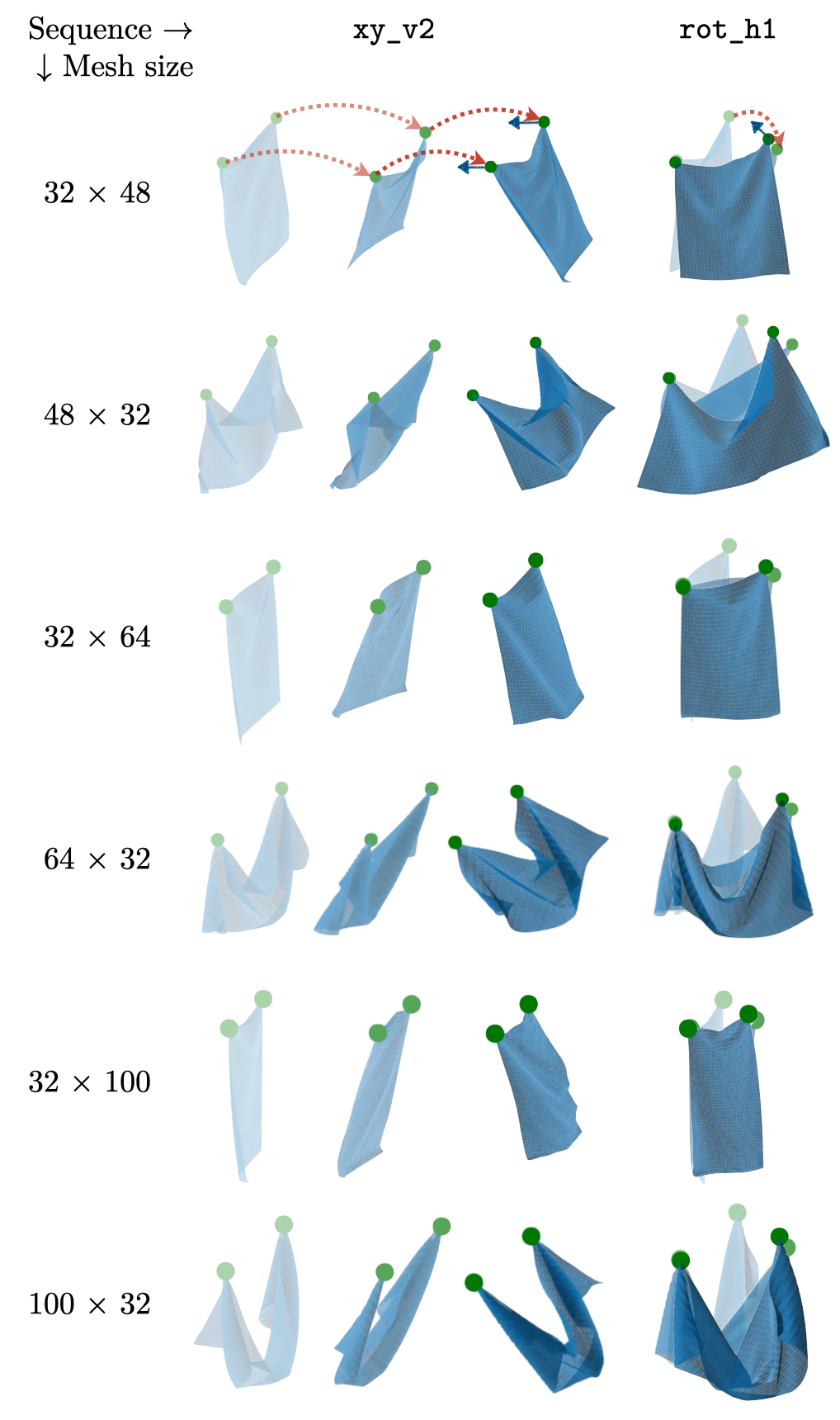}
   \caption{Generalization to non-square cloth meshes. 
}
   \label{fig:non_squared_mesh}
\end{figure}

\subsection{Ablation Study} \label{sec:supp_ablation}
\subsubsection{Number of iterations per time step}

\begin{figure}[H]
  \centering
\includegraphics[width=1\linewidth]{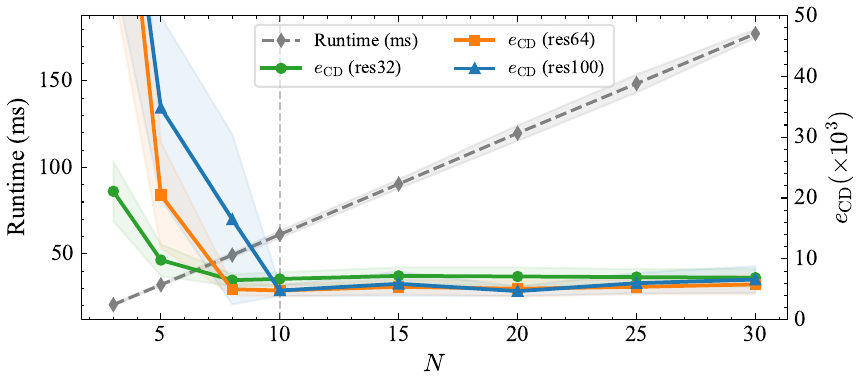}
\caption{\textsc{FNOpt} runtime (dashed gray line, left axis) and Chamfer distance $e_{\text{CD}}$ (solid colored lines, right axis) as functions of the number of iterations per time step. Green, orange, and blue curves correspond to mesh resolutions of $32\times32$, $64\times64$, and $100\times100$, respectively. Shaded bands denote one standard deviation over test sequences.}
   \label{fig:abl_iteration_vs_runtime}
\end{figure}

\subsection{Boundary Condition Generalization} \label{sec:supp_boundary_condition}
\paragraph{Non-square meshes.} Trained on square grid of the cloth, \textsc{FNOpt} is able to perform rollouts on grids with non-square aspect ratios. We demonstrate rollouts on six different mesh sizes across two motion sequences in \Cref{fig:non_squared_mesh}. Without any retraining or parameter tuning, the results indicate that the learned optimizer transfers across domain shapes, preserving fine-scale wrinkles and long-horizon stability.

\renewcommand{\thefootnote}{\fnsymbol{footnote}}

\subsubsection{FNO Architecture}
\begin{figure}[htbp]
  \centering
\includegraphics[width=1\linewidth]{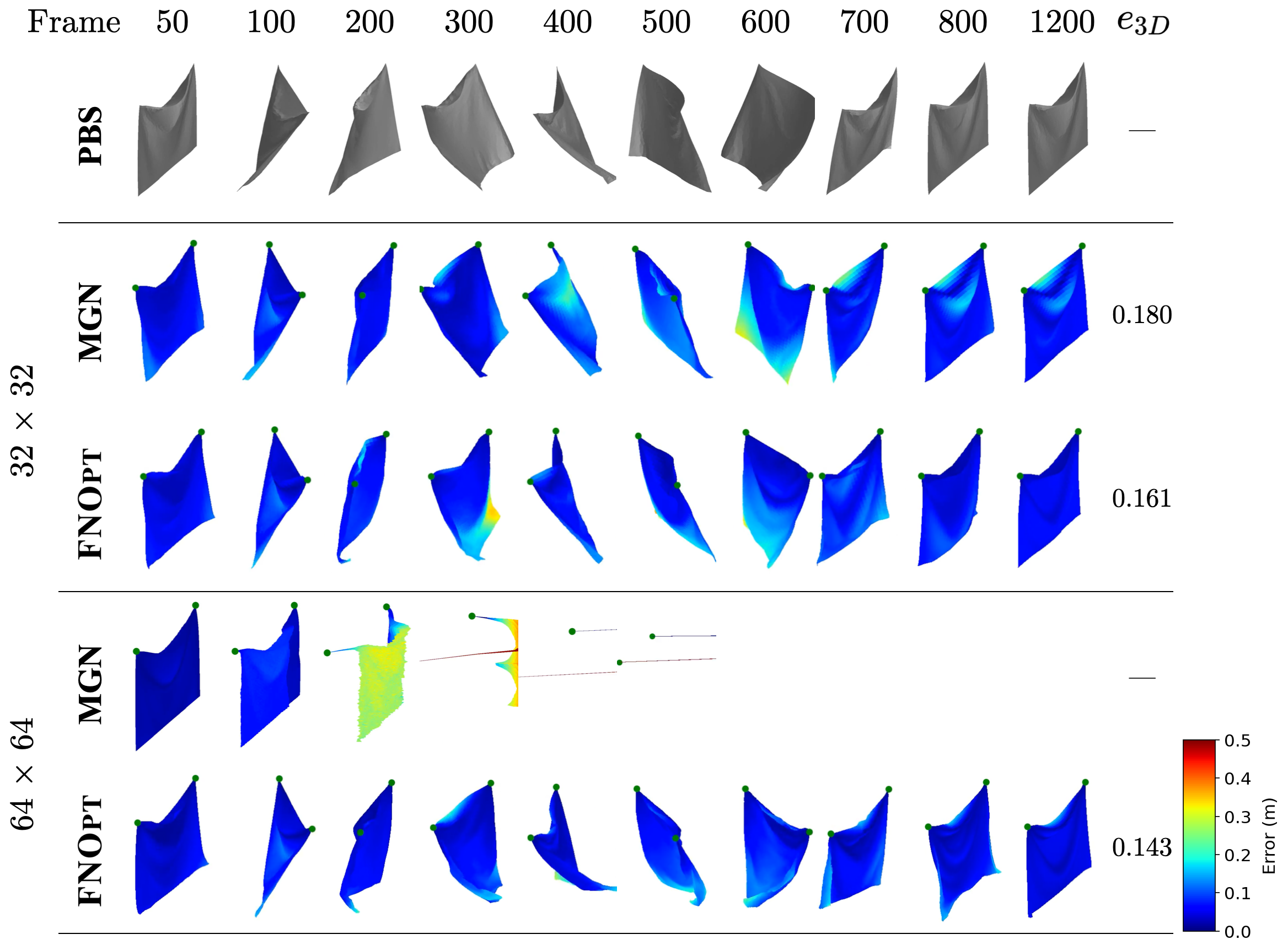}
   \caption{Vertex-wise error maps on the \texttt{rot\_h0} sequence, compared to MGN. Colors indicate the Euclidean distance between rollout and the PBS.}
   \label{fig:abl_mgn_error_map}
\end{figure}

We embed MeshGraphNets~\cite{Pfaff21} into the same meta-learning framework, while keeping the losses and training schedule unchanged. We refer to this variant as MGN. For hyper-parameters, we keep the default number of message-passing steps (15); the hidden size and the number of layers for the encoder, graph-net blocks, and decoder are set to 128 and 3, respectively. We train the network for $100$ epochs, which takes around 40 hours on an NVIDIA H100 GPU. We evaluate on the evaluation sequences of the datasets from \cite{Ian2023}; results are shown in \Cref{fig:abl_mgn_error_map}. The MGN variant exhibits visible high‑frequency oscillations already at the training resolution and diverges rapidly on $64 \times 64$ meshes.

\subsubsection{Neural Optimizer} \label{sec:supp_ablation_neural_optimizer}
\begin{figure}[htbp]
  \centering
\includegraphics[width=0.8\linewidth]{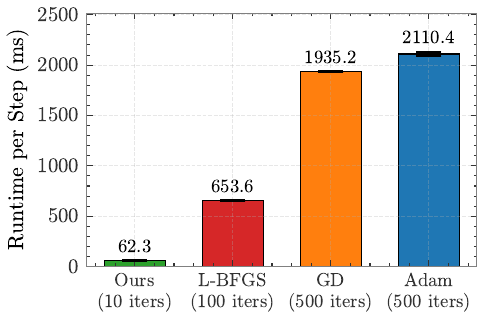}
\caption{Runtime comparison across different optimizers. Each bar shows per‑step runtime (ms); for every optimizer we use the minimum number of iterations that achieves convergence of $\mathcal{L}_{\mathrm{cloth}}$ (except GD, which did not converge). Results on $64\times64$ resolution. }
   \label{fig:abl_optimizer_runtime_comparison}

\end{figure}

We have compared \textsc{FNOpt} with classical first-order optimizers under varying numbers of inner iterations per time step: (a) vanilla gradient descent (GD) and (b) Adam. We additionally incorporate comparison with a quasi-Newton method, (c) limited-memory BFGS (L-BFGS). For each optimizer we run an independent learning rate sweep and report the best performing accuracy in \Cref{tab:abl_optimizers}. We report runtime in \Cref{fig:abl_optimizer_runtime_comparison}. All experiments were run on a $64\times64$ mesh. 

Even with a carefully tuned step size, GD fails to drive the physics loss $\mathcal{L}_{\mathrm{cloth}}$ to a low value with even $N=500$ iterations per time step, and the resulting roll‑outs exhibit visible artifacts. 

Thanks to its adaptive step size, Adam converges if a sufficiently large number of inner iterations (roughly $500$ per time‑step) is allowed. We use standard hyperparameters $\beta_1 = 0.9$ and $\beta_2 = 0.999$. However, its final $e_{\text{CD}}$ and $e_\text{3D}$ are still higher than ours, while being $30 \times$ slower.

For L‑BFGS, we compute the search direction via the standard two-loop recursion. To equalize gradient evaluations with first‑order baselines, we adopt a constant step size
and use a history size of $m{=}5$; we refer to this baseline as L‑BFGS (fixed step, no line search). This variant has $O(N^2)$ time complexity because of iteration over history, where $N$ is the number of iterations per time step, but it turns out to converge in fewer iterations than GD and Adam and achieves the most accurate results with $100$ iterations per time step. Yet, this is still $10\times$ slower than ours. Curiously, we evaluate its performance on a finer $100\times 100$ mesh with $70$, $100$, $120$, $150$ iterations per time step and record the $e_{\text{CD}}$ in \Cref{tab:abl_optimizers_lbfgs}. We find that it achieves the lowest metric at around 120 iterations per time step, and remains at low value for higher $N$. \textsc{FNOpt} achieves on par performance with only 10 iterations per time step.  

\begin{table}[t]
\centering
\setlength{\tabcolsep}{4pt}
\begin{tabular}{l| cccc| c}
\toprule
\multirow{3}{*}{\makecell{Sequence\\name}} &
\multicolumn{4}{c|}{L-BFGS (step size = 1)} &
\multicolumn{1}{c}{Ours} \\
\cmidrule(lr){2-5} \cmidrule(lr){6-6}
& \multicolumn{1}{c}{$70$} & \multicolumn{1}{c}{$100$} & \multicolumn{1}{c}{$120$} & \multicolumn{1}{c|}{$150$} & \multicolumn{1}{c}{$10$} \\
\midrule
\texttt{xy\_v2}       & 73.8 & 7.9 & 5.3 & 7.3 & 4.2 \\
\texttt{xy\_v2\_opp}  & 74.9 & 7.3 & 5.0 & 7.2 & 4.8 \\
\texttt{yz\_v2}       & 44.4 & 7.5 & 3.2 & 2.4 & 5.5 \\
\texttt{yz\_v2\_opp}  & 40.9 & 7.3 & 3.3 & 2.4 & 3.6 \\
\texttt{xz\_v2}       & 40.6 & 4.9 & 5.7 & 8.1 & 3.9 \\
\texttt{xyz\_v2}      & 91.1 & 6.6 & 4.0 & 5.0 & 4.3 \\
\texttt{xyz\_v2\_opp} & 87.7 & 6.4 & 3.9 & 5.2 & 5.2 \\
\texttt{xyz\_v3}      & 85.7 & 6.7 & 3.8 & 5.1 & 3.7 \\
\texttt{xyz\_v3\_opp} & 88.1 & 6.0 & 3.8 & 5.3 & 4.6 \\
\texttt{xyz\_v4}      & 90.5 & 6.9 & 3.8 & 5.1 & 3.5 \\
\texttt{xyz\_v4\_opp} & 89.5 & 6.9 & 3.8 & 4.8 & 4.7 \\
\texttt{rot\_h0}      & 27.1 & 5.1 & 5.3 & 8.0 & 6.2 \\
\texttt{rot\_h0\_opp} & 27.3 & 5.0 & 5.6 & 7.9 & 5.9 \\
\texttt{rot\_h1}      & 26.9 & 4.9 & 5.4 & 7.8 & 6.0 \\
\texttt{rot\_h1\_opp} & 26.8 & 5.1 & 5.8 & 7.9 & 6.2 \\
\midrule
\rowcolor[gray]{0.92}
Avg     & 58.3 & 6.1 & 4.7 & 5.8 & 4.7 \\
\rowcolor[gray]{0.92}
$\sigma$ & \std{25.2} & \std{1.1} & \std{0.9} & \std{1.7} & \std{0.9} \\

\bottomrule
\end{tabular}
\caption{Ablation on a $100\times 100$ mesh comparing $e_{\text{CD}}$ ($\times 10^{3}$) for L\text{-}BFGS and ours under varying \emph{numbers of iterations} per time step.}
\label{tab:abl_optimizers_lbfgs}
\end{table}

\begin{figure}[htbp]
  \centering
\includegraphics[width=1\linewidth]{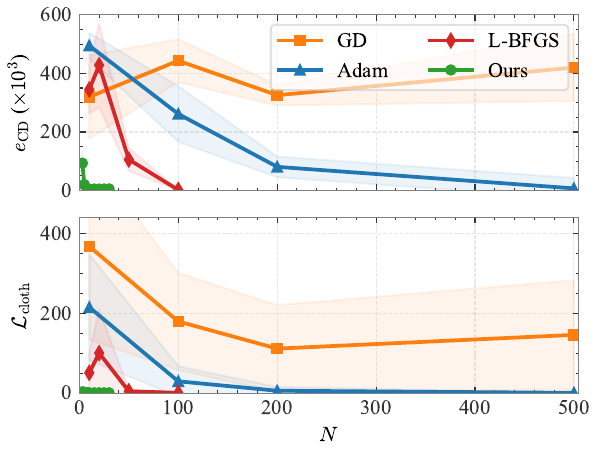}
\caption{Comparison of $e_{\text{CD}}$ and $\mathcal{L}_{\mathrm{cloth}}$ across classical optimizers and our method, versus the number of iterations per time step. Results on $64\times64$ resolution.}
   \label{fig:abl_optim_vs_iters}
\end{figure}

Finally, with only ten iterations per time step, \textsc{FNOpt} achieves low error and stable, high‑fidelity rollouts while being at least an order of magnitude faster than all classical optimizers. \Cref{fig:abl_optim_vs_iters} compares of $e_{\text{CD}}$ and $\mathcal{L}_{\mathrm{cloth}}$ across classical optimizers and our method as a function of $N$. Unlike classical optimizers, which require at least 100 iterations to achieve low error, \textsc{FNOpt} reaches both low $e_{\text{CD}}$ and $\mathcal{L}_{\mathrm{cloth}}$ within just 10 iterations.
\Cref{fig:abl_optimizer_runtime_comparison} shows the runtime comparison between different optimizers, evaluated on a node equipped with an AMD EPYC 7543 CPU and an NVIDIA A100 GPU.
This experiment highlights the advantage of a learned, resolution‑agnostic optimizer over classical update rules.

\subsubsection{Repulsive loss} \label{sec:supp_ablation_repulsive}
\begin{figure*}[t]
  \centering
\includegraphics[width=1\linewidth]{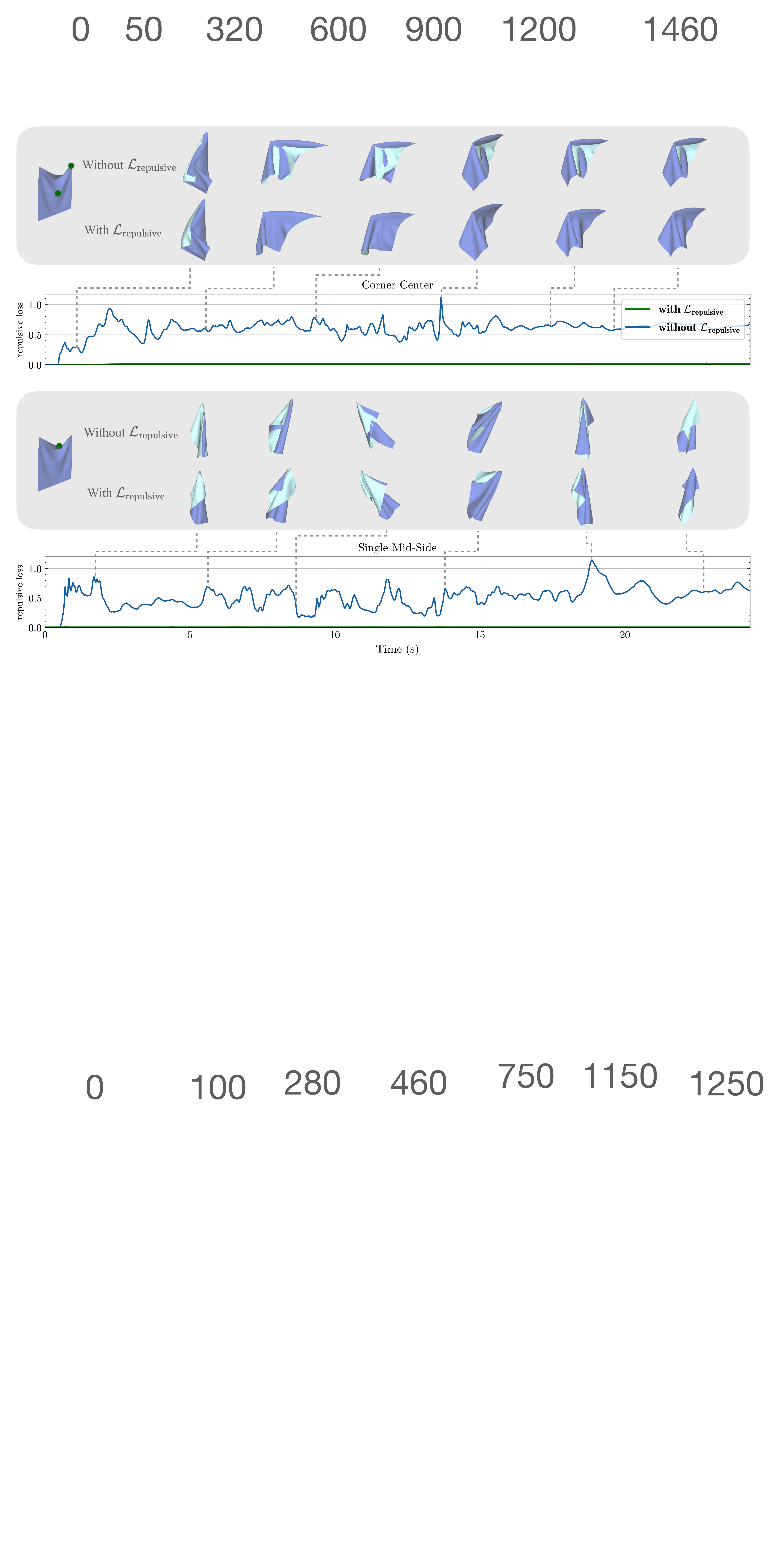}
\caption{Framewise repulsive loss with and without $\mathcal{L}_{\text{rep}}$. A subset of frames is shown for visual comparison: results without $\mathcal{L}_{\text{rep}}$ (left) versus with $\mathcal{L}_{\text{rep}}$ (right).}

\label{fig:abl_repulsive_framewise}

\end{figure*}

To alleviate self-collision, we applied a repulsive loss that penalizes non-adjacent cloth vertices when they are in close proximity. To assess its effectiveness, we consider two challenging scenarios with severe self-collision: a) Corner-center handle placement with motion sequence \texttt{rot\_h1}, and b) Single Mid-Edge with motion sequence \texttt{xy\_v2}. \Cref{tab:repulsive} reports the effect of the repulsive loss: the left column shows the average repulsive loss over the sequence; the middle column reports the proportions of frames with repulsive loss higher than 0.1, and the right column uses a stricter threshold of 0.02. \Cref{fig:abl_repulsive_framewise} further provides the framewise loss curves together with representative qualitative results, illustrating that the repulsive loss substantially reduces self-collision.  

\subsection{Simulation under Varying Material Coefficients}
\begin{figure}[htbp]
  \centering
\includegraphics[width=1\linewidth]{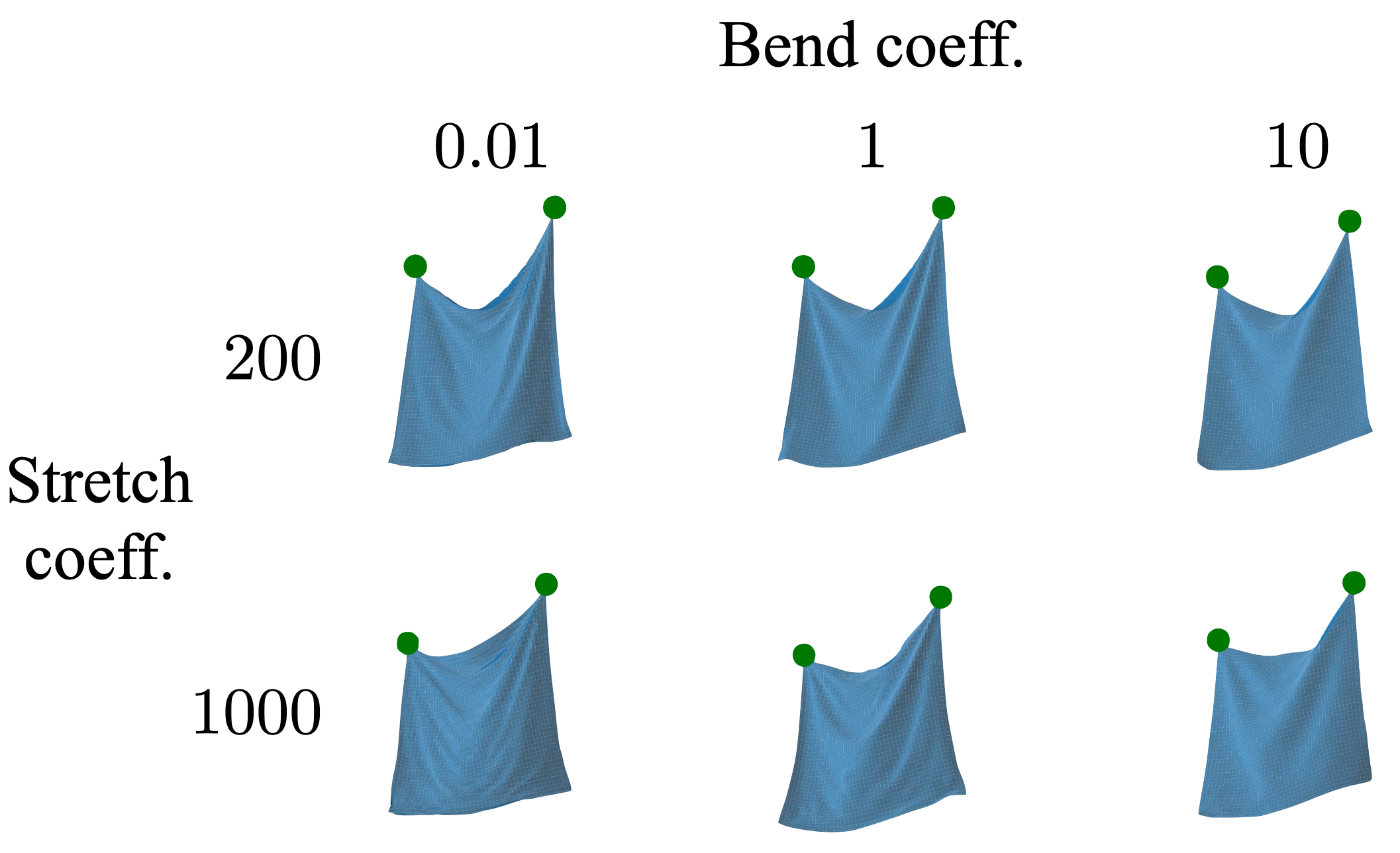}
\caption{Simulation results of selected frame with different stretching and bending coefficients.}
   \label{fig:material_coeff}

\end{figure}

\textsc{FNOpt} can simulate cloth with different stretching, shearing and bending coefficients. We visualize the results under varying stretching and bending coefficients in \Cref{fig:material_coeff}. The simulations are performed at $100 \times100$ resolution with $10$ iterations per time step.

\begin{table*}[t]
\centering
\setlength{\tabcolsep}{6pt}
\renewcommand{\arraystretch}{1.15}
\begin{tabular}{|l|ccc|ccc|}
\hline
 & \multicolumn{3}{c|}{Corner-Center} & \multicolumn{3}{c|}{Single Mid-Edge} \\
\cline{2-7}
 & $\mathcal{L}_{\text{rep}} (\times10^{3})\downarrow$
 & \% (0.1) $\downarrow$
 & \% (0.02) $\downarrow$
 & $\mathcal{L}_{\text{rep}} (\times10^{3})\downarrow$
 & \% (0.1) $\downarrow$
 & \% (0.02) $\downarrow$ \\
\hline
with $\mathcal{L}_{\text{rep}}$ & 13.12 ± 4.37 & 0.00 &  0.07 & 0.11±0.26 &  0.00 & 0.00 \\
no $\mathcal{L}_{\text{rep}}$   & 598.53 ± 141.29 & 97.95  & 98.08  & 508.79 ± 182.38 & 97.60 & 97.74  \\
\hline
\end{tabular}
\caption{Ablation study on the repulsive loss. We report average repulsive loss (scaled by $10^3$ for readability) and the percentage of frames with loss exceeding thresholds (0.02, 0.1). }
\label{tab:repulsive}
\end{table*}

\begin{table*}[t]
\centering
\setlength{\tabcolsep}{4pt}

\begin{tabular}{l| rrrr| rrrr| rrrr| r}
\toprule
\multirow{3}{*}{\makecell{Sequence\\name}} &
\multicolumn{4}{c|}{GD ($\text{lr}=0.01$)} &
\multicolumn{4}{c|}{Adam ($\text{lr}=0.1$)} &
\multicolumn{4}{c|}{L-BFGS ($\text{step size} = 1$)} &
\multicolumn{1}{c}{Ours} \\

\cmidrule(lr){2-5} \cmidrule(lr){6-9} \cmidrule(lr){10-13} \cmidrule(lr){14-14}
& \multicolumn{1}{c}{$10$} & \multicolumn{1}{c}{$100$} & \multicolumn{1}{c}{$200$} & \multicolumn{1}{c|}{$500$}
& \multicolumn{1}{c}{$10$} & \multicolumn{1}{c}{$100$} & \multicolumn{1}{c}{$200$} & \multicolumn{1}{c|}{$500$}
& \multicolumn{1}{c}{$10$} & \multicolumn{1}{c}{$20$} & \multicolumn{1}{c}{$50$} & \multicolumn{1}{c|}{$100$}
& \multicolumn{1}{c}{$10$}
\\\midrule
\texttt{xy\_v2}       & 390.2 & 492.8   &313.3 & 463.0 & 479.1 & 338.1 &  96.8 &  9.4 & 405.1 & 580.2 & 120.2 & 3.1 & 5.2 \\
\texttt{xy\_v2\_opp}  & 389.4 & 499.6   &316.3 & 466.8 & 480.0 & 320.4 & 115.6 &  7.7 & 400.3 & 579.5 & 119.3 & 3.2 & 5.3 \\
\texttt{yz\_v2}       & 491.4 & 434.4   &317.0 & 493.2 & 550.8 & 314.7 & 125.8 & 11.7 & 380.1 & 443.0 &  99.0 & 2.3 & 3.4 \\
\texttt{yz\_v2\_opp}  & 130.0 & 247.5   &213.0 & 401.5 & 454.7 & 282.6 & 105.5 &  8.0 & 270.0 & 438.3 &  89.7 & 2.4 & 3.2 \\
\texttt{xz\_v2}       & 282.9 & 417.9   &364.0 & 475.9 & 527.1 & 224.8 &   5.8 &  6.8 & 435.0 & 553.7 & 126.0 & 2.5 & 6.6 \\
\texttt{xyz\_v2}      & 540.6 & 507.0   &352.6 & 568.5 & 561.7 & 377.6 & 101.4 & 13.9 & 450.1 & 511.1 & 150.6 & 1.9 & 4.9 \\
\texttt{xyz\_v2\_opp} & 248.7 & 366.9   &309.1 & 435.1 & 453.7 & 304.1 & 116.5 & 10.6 & 350.6 & 500.5 & 140.8 & 2.5 & 4.1 \\
\texttt{xyz\_v3}      & 251.2 & 367.6   &309.3 & 433.8 & 448.4 & 327.7 &  98.6 &  5.1 & 350.8 & 501.3 & 134.9 & 2.5 & 4.7 \\
\texttt{xyz\_v3\_opp} & 535.0 & 505.7   &353.0 & 566.4 & 585.4 & 355.0 & 107.6 &  5.4 & 430.2 & 505.5 & 147.7 & 2.8 & 4.8 \\
\texttt{xyz\_v4}      & 536.0 & 504.7   &354.8 & 566.9 & 537.7 & 291.8 &  91.1 &  5.2 & 424.2 & 508.1 & 145.2 & 2.0 & 3.7 \\
\texttt{xyz\_v4\_opp} & 251.3 & 367.2   &313.3 & 433.9 & 446.6 & 327.7 &  81.3 &  5.4 & 363.1 & 499.8 & 132.0 & 3.1 & 4.2 \\
\texttt{rot\_h0}      & 180.5 & 483.9   &341.7 & 247.7 & 473.6 & 116.6 &  46.9 &  6.4 & 225.4 & 204.3 &  47.8 & 1.6 & 5.2 \\
\texttt{rot\_h0\_opp} & 193.2 & 486.5   &344.0 & 250.3 & 471.5 & 147.0 &  51.6 &  5.9 & 228.5 & 206.3 &  46.8 & 1.6 & 6.6 \\
\texttt{rot\_h1}      & 180.6 & 481.4   &340.2 & 248.7 & 463.2 &  91.8 &  42.8 &  8.2 & 223.4 & 186.1 &  47.1 & 1.7 & 4.5 \\
\texttt{rot\_h1\_opp} & 192.0 & 485.2   &342.6 & 246.7 & 482.2 & 103.7 &  29.8 &  5.4 & 235.7 & 199.3 &  46.2 & 1.7 & 5.7 \\
\midrule
\rowcolor[gray]{0.92}
Avg & 319.5 & 443.1 & 325.6 & 419.9 & 494.4 & 261.5 & 81.1 & 7.7 & 344.8 & 427.8 & 106.2 & 2.3 & 4.8  \\

\rowcolor[gray]{0.92}

$\sigma$ & \std{142.4} & \std{73.3} & \std{35.2} & \std{114.4} & \std{44.0} & \std{95.0} & \std{35.2} & \std{2.6} & \std{82.1} & \std{143.1} & \std{39.2} & \std{0.5} & \std{1.0} \\

\bottomrule
\end{tabular}

\caption{Ablation study on a $64\times64$ mesh comparing $e_{\text{CD}}$ ($\times 10^{3}$; lower is better) for different optimizers (GD, Adam, L-BFGS and ours) under varying \textit{numbers of iterations} per time‑step. 
}
\label{tab:abl_optimizers}
\end{table*}

\begin{table*}[t]
\centering
\setlength{\tabcolsep}{4pt}
\begin{tabular}{c|ccc|ccc}
\toprule
\multirow{3}{*}{$N$} 
& \multicolumn{3}{c|}{Runtime (ms)} 
& \multicolumn{3}{c}{$e_{\text{CD}}$ ↓} \\
\cmidrule(lr){2-4} \cmidrule(lr){5-7}
& res32 & res64 & res100 
& res32 & res64 & res100 \\
\midrule
$3$  & \avgstd{20.6}{0.7} & \avgstd{20.6}{1.1} & \avgstd{21.9}{0.9} & \avgstd{21.1}{4.9} & \avgstd{93.7}{38.9} & \avgstd{93.1}{35.9} \\
$5$  & \avgstd{32.1}{0.7} & \avgstd{32.2}{1.1} & \avgstd{32.4}{0.4} & \avgstd{9.8}{2.6} & \avgstd{20.5}{8.7} & \avgstd{34.8}{15.0} \\
$8$  & \avgstd{49.3}{1.3} & \avgstd{52.3}{2.2} & \avgstd{50.8}{4.9} & \avgstd{6.5}{1.0} & \avgstd{5.0}{1.0} & \avgstd{16.5}{14.0} \\
$10$ & \avgstd{61.2}{2.1} & \avgstd{62.3}{2.3} & \avgstd{61.7}{1.3} & \avgstd{6.7}{1.1} & \avgstd{4.8}{1.0} & \avgstd{4.7}{0.9} \\
$15$ & \avgstd{90.4}{2.7} & \avgstd{91.7}{3.2} & \avgstd{90.8}{1.3} & \avgstd{7.2}{1.4} & \avgstd{5.4}{1.2} & \avgstd{5.9}{2.0} \\
$20$ & \avgstd{119.8}{4.5} & \avgstd{119.3}{2.2} & \avgstd{120.4}{3.6} & \avgstd{7.1}{1.3} & \avgstd{5.1}{1.3} & \avgstd{4.7}{0.8} \\
$25$ & \avgstd{148.5}{5.2} & \avgstd{147.3}{1.8} & \avgstd{149.6}{1.8} & \avgstd{7.0}{1.3} & \avgstd{5.4}{1.2} & \avgstd{6.0}{1.7} \\
$30$ & \avgstd{177.4}{2.5} & \avgstd{177.5}{3.5} & \avgstd{178.7}{3.6} & \avgstd{6.9}{1.5} & \avgstd{5.8}{1.6} & \avgstd{6.6}{2.2} \\

\bottomrule
\end{tabular}
\caption{Ablation study with $N\in \{3,5,8,10,15,20,25,30\}$ iterations per time‑step across three mesh resolutions. We report the per-frame runtime (ms) and $e_{\text{CD}}$ ($\times 10^{3}$).}\label{tab:abl_n_iteration_time}
\end{table*}

\begin{table*}[!t]
\centering
\setlength{\tabcolsep}{4pt}

\begin{tabular}{l@{\hskip 2pt}r@{\hskip 3pt}r@{\hskip 3pt}r@{\hskip 3pt}r@{\hskip 3pt}r@{\hskip 3pt}r@{\hskip 3pt}r@{\hskip 3pt}r@{\hskip 3pt}r}

\toprule
\multirow{3}{*}{\makecell{Sequence\\name}} &
\multicolumn{3}{c}{$32\times32$} &
\multicolumn{3}{c}{$64\times64$} &
\multicolumn{3}{c}{$100\times100$} \\

\cmidrule(lr){2-4} \cmidrule(lr){5-7} \cmidrule(lr){8-10}
& \multicolumn{1}{c}{5 iters} & \multicolumn{1}{c}{10 iters} & \multicolumn{1}{c}{20 iters}
& \multicolumn{1}{c}{5 iters} & \multicolumn{1}{c}{10 iters} & \multicolumn{1}{c}{20 iters}
& \multicolumn{1}{c}{5 iters} & \multicolumn{1}{c}{10 iters} & \multicolumn{1}{c}{20 iters}
\\\midrule
\texttt{xy\_v2}       & \cfd{ 6.2}{16.1} & \cfd {7.0}{15.2} &\cfd{7.6}{15.5} & \cfd{19.2}{24.4} & \cfd{5.2}{12.7} & \cfd{5.2}{12.4} & \cfd{64.5}{46.5} & \cfd{4.2}{11.0} & \cfd{4.8}{10.5}  \\
\texttt{xy\_v2\_opp}  & \cfd{ 5.9}{14.7} & \cfd{ 6.5}{14.5} &\cfd{7.7}{14.9} & \cfd{24.7}{24.0} & \cfd{5.3}{12.0} & \cfd{5.6}{12.1} & \cfd{32.8}{30.5} & \cfd{4.8}{10.5} & \cfd{5.5}{10.9}  \\
\texttt{yz\_v2}       & \cfd{11.3}{21.2} & \cfd{ 4.9}{15.0} &\cfd{4.0}{13.9} & \cfd{ 8.3}{17.6} & \cfd{3.4}{11.3} & \cfd{3.3}{10.8} & \cfd{19.9}{24.3} & \cfd{5.5}{13.1} & \cfd{3.4}{10.8}  \\
\texttt{yz\_v2\_opp}  & \cfd{11.0}{19.4} & \cfd{ 4.3}{13.5} &\cfd{5.2}{13.8} & \cfd{ 6.4}{15.1} & \cfd{3.2}{10.3} & \cfd{2.9}{09.8} & \cfd{38.7}{27.7} & \cfd{3.6}{10.9} & \cfd{3.6}{09.5}  \\
\texttt{xz\_v2}       & \cfd{16.6}{21.6} & \cfd{ 8.9}{15.2} &\cfd{8.6}{14.9} & \cfd{21.3}{22.7} & \cfd{6.6}{13.3} & \cfd{6.1}{10.7} & \cfd{24.3}{23.6} & \cfd{3.9}{ 9.9} & \cfd{5.4}{10.3}  \\
\texttt{xyz\_v2}      & \cfd{10.0}{18.5} & \cfd{ 6.4}{14.6} &\cfd{6.4}{14.1} & \cfd{14.0}{20.5} & \cfd{4.9}{12.0} & \cfd{4.3}{10.5} & \cfd{44.4}{31.0} & \cfd{4.3}{10.5} & \cfd{4.0}{ 8.6}  \\
\texttt{xyz\_v2\_opp} & \cfd{ 9.1}{17.7} & \cfd{ 6.4}{14.7} &\cfd{5.3}{13.1} & \cfd{16.5}{18.6} & \cfd{4.1}{10.8} & \cfd{4.5}{12.7} & \cfd{27.4}{27.0} & \cfd{5.2}{11.4} & \cfd{4.4}{10.9}  \\
\texttt{xyz\_v3}      & \cfd{12.0}{19.4} & \cfd{ 6.4}{15.2} &\cfd{5.7}{13.2} & \cfd{27.6}{24.4} & \cfd{4.7}{12.2} & \cfd{4.4}{12.1} & \cfd{41.1}{31.8} & \cfd{3.7}{10.6} & \cfd{4.3}{10.7}  \\
\texttt{xyz\_v3\_opp} & \cfd{12.1}{19.6} & \cfd{ 6.1}{14.3} &\cfd{6.9}{14.8} & \cfd{18.8}{21.3} & \cfd{4.8}{11.8} & \cfd{4.5}{10.7} & \cfd{27.4}{23.8} & \cfd{4.6}{10.9} & \cfd{5.0}{10.0}  \\
\texttt{xyz\_v4}      & \cfd{ 6.5}{15.8} & \cfd{ 6.7}{14.4} &\cfd{8.3}{15.5} & \cfd{21.4}{22.7} & \cfd{3.7}{10.7} & \cfd{4.8}{11.9} & \cfd{29.5}{29.4} & \cfd{3.5}{ 9.9} & \cfd{4.0}{10.0}  \\
\texttt{xyz\_v4\_opp} & \cfd{ 8.7}{17.5} & \cfd{ 7.4}{15.2} &\cfd{7.2}{14.7} & \cfd{25.3}{22.9} & \cfd{4.2}{11.2} & \cfd{4.0}{10.9} & \cfd{31.1}{31.8} & \cfd{4.7}{11.2} & \cfd{4.1}{10.5}  \\
\texttt{rot\_h0}      & \cfd{ 9.1}{20.4} & \cfd{ 6.9}{16.1} &\cfd{7.7}{16.8} & \cfd{14.3}{22.1} & \cfd{5.2}{14.3} & \cfd{7.5}{15.8} & \cfd{18.3}{26.3} & \cfd{6.2}{15.1} & \cfd{4.6}{12.1}  \\
\texttt{rot\_h0\_opp} & \cfd{ 8.5}{19.6} & \cfd{ 7.6}{17.1} &\cfd{9.5}{18.3} & \cfd{33.3}{22.3} & \cfd{6.6}{15.8} & \cfd{6.8}{14.2} & \cfd{16.7}{25.0} & \cfd{5.9}{14.8} & \cfd{5.8}{13.3}  \\
\texttt{rot\_h1}      & \cfd{10.2}{21.0} & \cfd{ 7.6}{16.8} &\cfd{8.4}{17.5} & \cfd{40.8}{31.1} & \cfd{4.5}{13.9} & \cfd{6.2}{14.5} & \cfd{70.4}{54.3} & \cfd{6.0}{14.6} & \cfd{5.1}{13.2}  \\
\texttt{rot\_h1\_opp} & \cfd{ 9.2}{21.1} & \cfd{ 7.5}{17.0} &\cfd{8.7}{17.9} & \cfd{16.3}{25.9} & \cfd{5.7}{15.8} & \cfd{7.1}{15.7} & \cfd{36.1}{34.4} & \cfd{6.2}{14.7} & \cfd{6.0}{13.5}  \\
\midrule
Avg                   & \cfd{ 9.8}{0.189} & \cfd{ 6.7}{0.153} &\cfd{7.2}{0.153} & \cfd{20.5}{0.224}& \cfd{4.8}{0.126} & \cfd{5.1}{0.124} & \cfd{34.8}{0.312} & \cfd{4.7}{0.119} & \cfd{4.7}{0.110}  \\
\addlinespace[-0.5ex]
 
 $\sigma$ & \stdp{2.6}{0.021} & \stdp{1.1}{0.010} & \stdp{1.5}{0.016} & \stdp{8.7}{0.036} & \stdp{1.0}{0.017} &\stdp{1.3}{0.019} & \stdp{15.0}{0.083} & \stdp{0.9}{0.019} & \stdp{0.8}{0.014} \\
\bottomrule
\end{tabular}
\caption{Per-sequence $e_{\text{CD}}$ ($\times 10^{3}$) and $e_\text{3D}$ error ($\times 10^{2}$; values in parentheses) at \(5\), \(10\), and \(20\) iteration budgets, evaluated on three mesh resolutions.}\label{tab:abl_n_iteration}
\end{table*}

\clearpage